\documentclass{article} 
\usepackage{iclr2025_conference,times}
\usepackage{natbib}
\usepackage[utf8]{inputenc}

\usepackage{amsmath,amsfonts,bm}

\def\eqref#1{equation~\ref{#1}}

\def\1{\bm{1}}

\DeclareMathAlphabet{\mathsfit}{\encodingdefault}{\sfdefault}{m}{sl}
\SetMathAlphabet{\mathsfit}{bold}{\encodingdefault}{\sfdefault}{bx}{n}

\usepackage{hyperref}
\usepackage{url}
\usepackage{wrapfig}
\usepackage{comment}

\usepackage{amssymb}
\usepackage{graphicx}
\usepackage{array}
\usepackage{ragged2e}

\usepackage{booktabs}

\usepackage{makecell}

\usepackage{listings}
\usepackage{multirow}

\usepackage[inline]{enumitem}

\title{What I cannot execute, I do not understand: Training and Evaluating LLMs on Program Execution Traces}

\author{Jordi Armengol-Estapé$^1$\thanks{Work done while interning at FAIR, Meta AI. Contact: \texttt{jordi.armengol.estape@ed.ac.uk}}\ \ , Quentin Carbonneaux$^2$, Tianjun Zhang$^2$, \\ \textbf{Aram H. Markosyan$^2$, Volker Seeker$^2$, Chris Cummins$^2$, Melanie Kambadur$^2$,} \\
\textbf{Michael F.P. O'Boyle$^1$, Sida Wang$^2$, Gabriel Synnaeve$^2$, Hugh Leather$^2$} \\
$^1$University of Edinburgh  \quad
$^2$Meta AI
}

\newcommand{\name}{E.T.}

\newcommand{\fullname}{Execution Tuning} 

\iclrfinalcopy 
\begin{document}

\maketitle

\begin{abstract}
Code generation and understanding are critical capabilities for large language models (LLMs). Thus, most LLMs are pretrained and fine-tuned on code data. However, these datasets typically treat code as static strings and rarely exploit the dynamic information about their execution. Building upon previous work on trace modeling,  we study \fullname{} (\name{}), a training procedure in which we explicitly model real-world program execution traces without requiring manual test annotations. 
We train and evaluate models on different execution trace granularities (line and instruction-level) and strategies on the task of output prediction, obtaining ${\sim}80\%$ accuracy on CruxEval and MBPP, and showing the advantages of \textit{dynamic scratchpads} (i.e., self-contained intermediate computations updated by the model rather than accumulated as a history of past computations)  on long executions (up to 14k steps). Finally, we discuss \name{}'s practical applications.
\end{abstract}

\section{Introduction}
\label{sec:intro}

Coding capabilities are one of the most important applications of large language models (LLMs) \citep{NEURIPS2020_1457c0d6}, for which LLMs specialized on coding have been trained on large-scale datasets of programming languages \citep{chen2021codex, roziere2024codellamaopenfoundation}. Current state-of-the-art general-purpose LLMs are thought to contain considerable proportions of code in their pretraining data \citep{openai2024gpt4technicalreport},
which is known to improve reasoning capabilities even in tasks seemingly unrelated to code \citep{aryabumi2024codecodeexploringimpact}.

However, datasets used to train code LLMs (such as \citet{lozhkov2024starcoder2stackv2}) typically treat code as static strings and rarely exploit the \textit{dynamic} information about their execution. Executability is one of the key differences between code and natural language, and most code datasets neglect dimensions of the code domain such as reasoning over code execution, which in turn could lead to better code understanding. 
 
This fundamental limitation has sparked a renewed interest in modeling program executions, connecting with the pre-LLM neural program evaluation literature \citep{learning_to_execute2014, GravesWD14}, which studied whether neural networks could learn to execute programs. \citet{Austin2021} fine-tune LLMs to directly predict the output of Python functions from coding competitions and math problems, which are paired with unit tests.
Crucially, \citet{scratchpad} showed that asking (and training) the model to predict all the line-level states of a Python function execution up to the return value improved the results on function output prediction, compared to directly asking to predict the return value. They refer to these tokens emitted by the model to perform intermediate computations before the final answer as \textit{scratchpad}. In this work, we build upon this approach. 

Nevertheless, key questions remain unanswered:
\begin{enumerate*} 
\item How we increase the number of examples in trace datasets, given that the programs need to be executable?
\item How does trace granularity affect the models's performance?
\item How can we handle long execution traces?
\item What kind of scratchpad works best for storing intermediate outputs - can we skip ``unnecessary'' intermediate steps?
\item What are the effects of trace modeling on downstream code generation tasks?
\end{enumerate*}

\begin{figure}[thbp!]
\includegraphics[width=\textwidth]{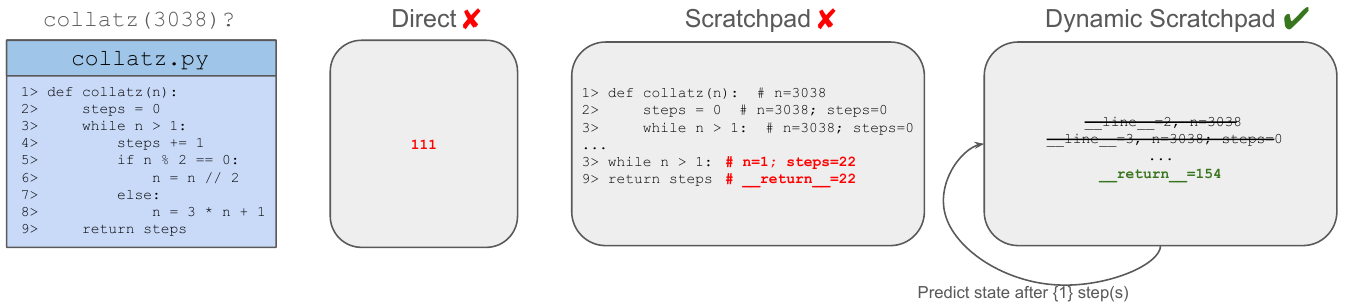}
  
    \caption{Given a natural number, a function returns the number of iterations required to arrive at 1, when following the sequence in the Collatz conjecture. Can we predict the output of such a function for large inputs (3038 in our example) using LLMs? Asking an LLM to directly predict the output results in a plausible but incorrect answer.  
    Training a model to predict the intermediate traces of the function as a scratchpad of intermediate computations  \citep{scratchpad} generally yields more accurate output predictions, but can be impractical or even inaccurate with long executions.
    In this work, we introduce \textit{dynamic}  scratchpads, in which the model updates a single, self-contained scratchpad instance, 
    yielding to more accurate predictions for long executions. 
    }
    \label{fig:scratchpad}
\end{figure}

With the goal of answering these questions, we study \fullname{} (\name{}), a training procedure in which we explicitly
model real-world program execution traces without requiring manual test annotations (needed to execute the programs we want to trace). To scale trace modeling to large, real-world programs, we start from a collection of $\sim$300k Python functions, made executable with synthetic inputs generated by a combination of LLMs and fuzzing.
We then build a custom Python tracer to track local variables, global variables, and additional information obtained from the stack. We statically represent traces in LLM-friendly formats, including iterators and functions. After trace collection, to ingest traces to LLMs we study three levels of granularity: program (i.e., direct output prediction), line, and bytecode instructions.

We compare three scratchpad strategies for storing the intermediate computations: a) regular scratchpad \citep{scratchpad}, i.e., a dictionary with all the variable values at each step, b) \textit{compact} scratchpad containing the changed variables only \citep{ni2024nextteachinglargelanguage}, and c) \textit{dynamic} scratchpad (depicted in Figure \ref{fig:scratchpad}), in which, rather than accumulating all the intermediate computation history, the LLM is asked to update a single, self-contained representation of the current state of the program.

As a proxy of code reasoning, we evaluate models on program output prediction (given an input), allowing them to generate  intermediate execution states. We first evaluate on the standard output prediction benchmark, CruxEval \citep{gu2024cruxeval}, on which models trained on traces clearly outperform the direct output prediction ones. 
However, we also observe interesting failure modes involving indexing and basic string manipulation. Aiming at evaluating on longer and more diverse executions, we also run our models on a subset of a Python synthesis benchmark, MBPP \citep{mbpp}, selecting functions with nested loops, where we observe higher disparity between tracing strategies. To study even longer executions, we also study algorithmic tasks with arbitrarily long execution lengths, including the Collatz conjecture (also known as the Ulam conjecture), 
showing the advantages of dynamic scratchpads on long executions (success on up to 14k execution steps) and the potential of dynamically skipping steps (allowing to decrease the needed intermediate steps from e.g. 14k steps to 1.5k). Finally, we discuss applications by analyzing the effects of \name{} on code generation and reasoning tasks.

\section{Execution Tuning}
\label{sec:traces}

\begin{figure*}[thbp!]
\centering
\includegraphics[width=0.8\textwidth]{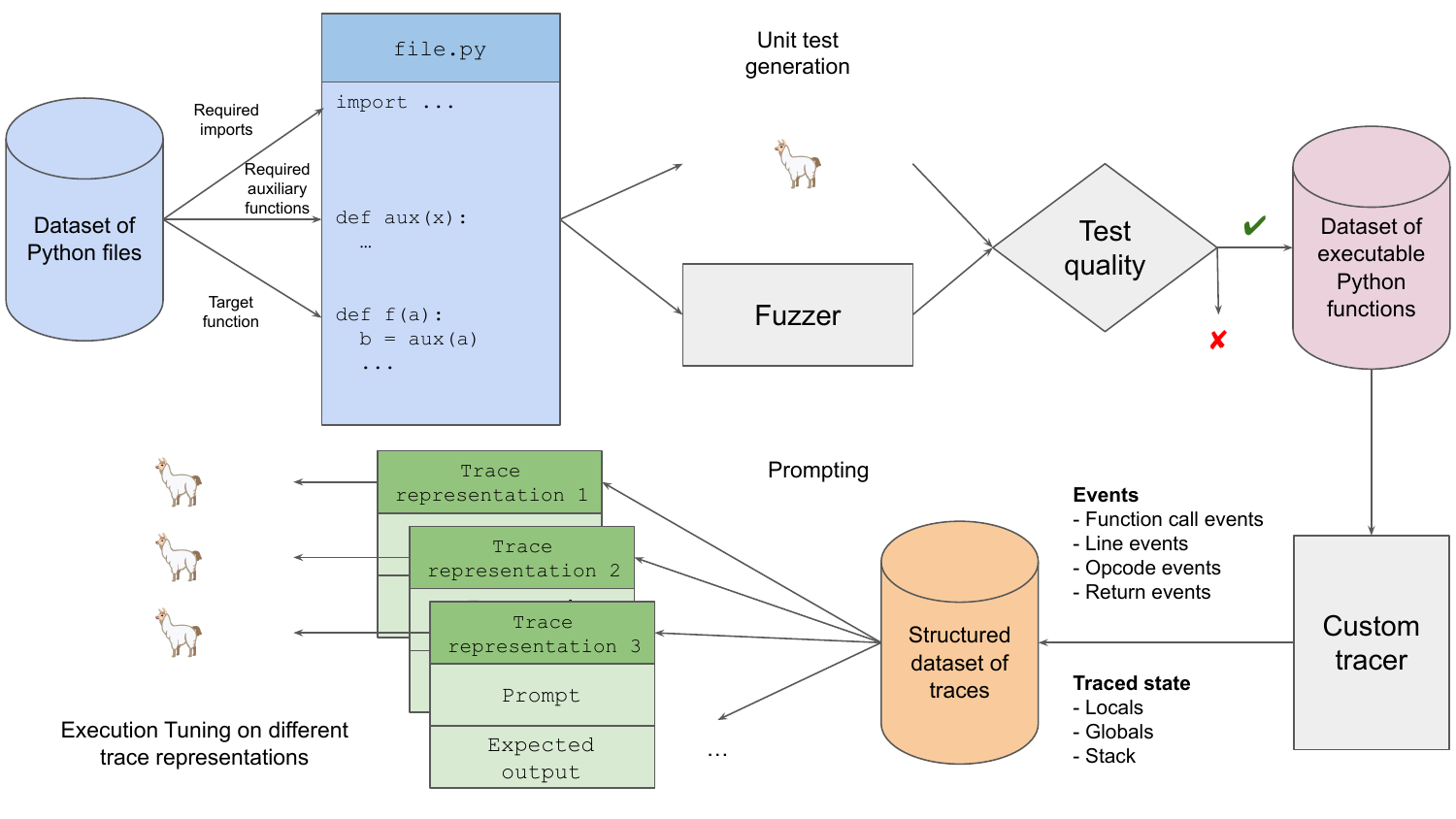}
    \caption{Overview of the data pipeline in \name{} We start from Python functions made executable with synthetic yet representative inputs generated by a combination of LLMs and fuzzing, filtered by test quality. Our custom Python tracer generates a structured dataset of traces. From this dataset, we train models prompted with different trace representations.}
    \label{fig:overview}
\end{figure*}

In this section, we describe the two implementation challenges of \name{} The first one is about where and how to collect execution traces to construct a large and representative training dataset. The second challenge is how to represent these traces to ingest them to the model. Figure \ref{fig:overview} shows an overview of the pipeline for these two challenges.

\subsection{Traces collection}

We start from a collection of unrestricted Python code from where functions are extracted, together with their corresponding module imports and auxiliary functions. We allow importing common Python libraries such as \texttt{pandas} or \texttt{matplotlib}. The inputs (function arguments) are generated with an LLM - more specifically, by prompting Llama 3 8B \citep{dubey2024llama3herdmodels} to generate unit tests for these functions. 
For increased coverage, inputs are also generated using fuzzing. In both cases, inputs yielding runtime errors are discarded, and inputs are filtered on test quality by measuring line coverage and similarity between tests.

In total, combining the LLM and fuzzing generated inputs,  we gather about $\sim$300k executable functions, with an average of 6 inputs per function. Using automatically generated inputs allows us to scale the training dataset to 
 $>$ 1.5M executions, without requiring manually written unit tests.
  \begin{wrapfigure}{r}{0.4\textwidth} 
    \centering
    \includegraphics[width=0.4\textwidth]{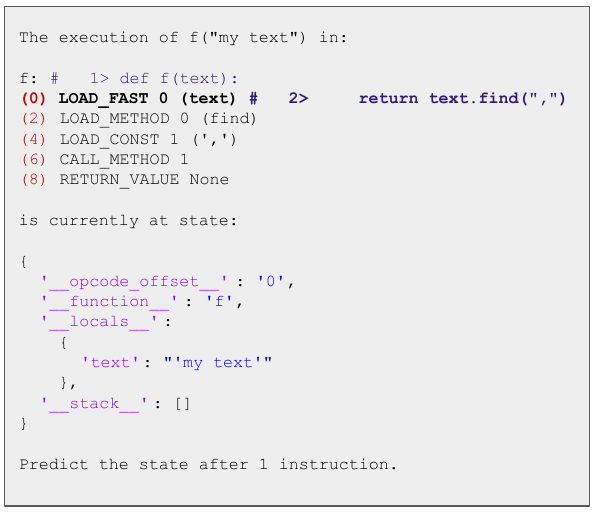}
    \caption{Prompt for Instruction-1.}
    \label{fig:ex}
\end{wrapfigure}

We build a custom tracer leveraging Python's built-in \texttt{sys.settrace}. We capture all Python function call, return, line  and opcode events, and step into user-defined auxiliary functions (but not into functions from imported modules). We deliberately  ignore C events because without access to the source code emitting these events, the C traces would introduce noise to the data. We note that for correctly discarding the 
 C traces, we need to explicitly deactivate tracing of Python code called from C (e.g. a C function calling a Python lambda).
 We step into auxiliary functions present in the context (i.e., defined in the same file) and but don't do so for functions out of the context. Unlike previous work \citep{scratchpad, ni2024nextteachinglargelanguage}, we also capture globals, the stack, and state changes at the instruction-level granularity (i.e. Python opcodes). 
After tracing, we have constructed a structured dataset of executions. The next step is to turn this structured dataset into concrete (prompt, expected output) pairs to ingest to the models.

\subsection{Traces representation}
Following \citet{ni2024nextteachinglargelanguage}, we rely on Python's \texttt{\_\_repr\_\_} 
to have an LLM-friendly representation of each object. Unlike in \citet{ni2024nextteachinglargelanguage}, where authors summarized loops with the first two and the last iterations,  here we want to represent complete program executions, for which we consider different strategies.

\textbf{Granularities}
We study the following trace granularities: \begin{enumerate*}

\item Direct predictions: Following \citet{Austin2021, gu2024cruxeval}, we fine-tune the models to directly predict the output (return value) from the input of the function.

\item Line-level (source code): Following \citet{scratchpad}, we represent the states at each executed line.

\item Instruction-level (bytecode): Source code lines can map to multiple instructions at the assembly or bytecode level. 
With this motivation in mind, we also consider a representation in which we explicitly show instructions and instruction-level state to the model. Crucially, this implies the introduction of the stack. Aside from the bytecode, the model has access to the source code, shown as inline comments at the first opcode of the corresponding line.
\end{enumerate*}

\textbf{Scratchpads}
We consider the following scratchpad techniques for storing intermediate computations (i.e., the traces):
\begin{enumerate*}
    \item Scratchpad: Following the original scratchpad work \citep{scratchpad}, the model predicts the state after executing each line, defined as the line itself plus the dictionary of the local variables, followed by the predicted return value.
    \item \textit{Compact} scratchpad: Inspired by \citet{ni2024nextteachinglargelanguage}'s trace representation, we also consider a \texttt{diff}-based scratchpad, in which the model only needs to predict the variables that change with respect  to the previous state. This should help at long executions by decreasing the token count. Note, though, that in \citet{ni2024nextteachinglargelanguage} this representation was not used as a scratchpad, but to annotate code.

    \item \textit{Dynamic} scratchpad: The two previous scratchpad strategies ask the model to predict the entire execution history of a program (paired with an input), up to the return value. This is problematic with long executions. With this motivation in mind, we introduce dynamic scratchpads, in which a single, self-contained state is updated by the model at every step. It also has the additional advantage that with the same strategy we can naturally train the model to skip steps that are potentially unnecessary to predict the final output, by asking the model to predict what the state will be after N steps. The caveat is that parts of the state that were implicitly encoded by having access to the execution history, now will need to be encoded explicitly. In particular, iterator states, not part of the locals dictionary in Python, can be ambiguous.\footnote{For example, in \texttt{for c in ('a','b','a')}, if we only have access to the current state, we need a way of distinguishing between the first \texttt{'a'} and the last one.
    We explicitly encode it with e.g., \texttt{\_\_for\_iterator\_1\_\_=2} lets us know the iteration count on a given iterator.
    } For this reason, even for the models with line-level granularity, we access the stack to trace the iterators, and explicitly encode their iteration count.
\end{enumerate*}

Figure \ref{fig:scratchpad} depicts the differences between direct output prediction, scratchpad-based output prediction, and dynamic scratchpad. Compact scratchpad is omitted for brevity; it's similar to scratchpad but just predicting the variables that change. Figure \ref{fig:ex} provides a prompt example.

\section{Output prediction results}
\label{sec:intrinsic}

\begin{table*}[thbp!]
\centering
\caption{\label{tab:intrinsic_crux_e}
Results of individual state predictions on CruxEval, i.e. before aggregating steps into full executions for output prediction. The accuracy is broken down into control flow (does the model  correctly predict the next line?), variables (does the model correctly predict the variable values in the next state?), iterators (does the model correctly predict the iteration count for the current iterators?) stack state, and full state accuracy (how many states are completely correct, i.e. control flow, variables, iterators, and stack are all correct, assuming the state had them). Note that scratchpad does not have iterator states because it does not require them, and line-level models do not have access to the stack. In this Table, F.T. means that the models were fine-tuned on the task, while prompted results indicate no training on traces. %
}
\scalebox{0.95}{
\begin{tabular}{llrrrrr}
\hline
\textsc{Repr.} & \textsc{Model} & \textsc{C. flow} & \textsc{Vars} & \textsc{Iterator} & \textsc{Stack} & \textsc{Full}\\
\hline

Scratchpad & Llama3.1 8B + F.T. & 91.9\% & 86.5\% & - & - & 86.4\%\\

\hline
Line-1  & Llama3.1 8B (prompted)  & 53.8\% & 26.9\% & 39.6\% & - & 10.6\%\\
& \hspace{3mm} + F.T. & 99.5\% & 97.7\% & 99.7\% & - & 96.3\%\\

\hline 
Line-n (global) & Llama3.1 8B (prompted)  & 16.8\% &16.0\%  & 12.3\% & -  & 1.8\%\\
 & \hspace{3mm} + F.T. & 95.1\% & 66.8\% & 96.4\% & - & 79.0\%\\

\hline 
Instruction-1 & Llama3.1 8B (prompted) & 74.1\% & 80.4\% & 77.9\%& 5.8\% & 2.8\% \\
 & \hspace{3mm} +F.T.  & 99.9\% & 99.9\% & 99.9\% & 98.8\% & 98.8\% \\

\hline
\end{tabular}
} 

\end{table*}

In this section, we evaluate models on function output prediction, a proxy for code reasoning, comparing different trace representation strategies. All evaluated models are  fine-tuned using comparable hyperparameters from the instruct version of Llama 3.1-8B \citep{dubey2024llama3herdmodels}, unless stated otherwise.

We start by analyzing the results for individual step predictions. Then, we aggregate these step predictions to evaluate output prediction on CruxEval. Next,  aiming at evaluating on longer and more diverse executions, we also run our models on a subset of MBPP (selecting functions with nested loops) and algorithmic tasks with arbitrarily long execution lengths.

We will refer to the dynamic scratchpad models by using \{Line$\mid$Instruction\}-\{1$\mid$n\}, where Line models have a granularity of lines and Instruction models have a granularity of bytecode instructions. ``1" refers to models trained to predict the next step, while ``n" refers to models trained to predict multiple steps into the future, with n = \{1..10\}. Additionally, note that in our results, we refer to our re-implementation of \citet{scratchpad} as ``Scratchpad", which benefits from the increased data size and context length. The original Scratchpad restricted context windows to 512 tokens; here, we allow up to 8192 tokens.

\begin{wrapfigure}{r}{0.5\textwidth}

    \centering

    \includegraphics[width=0.5\textwidth]{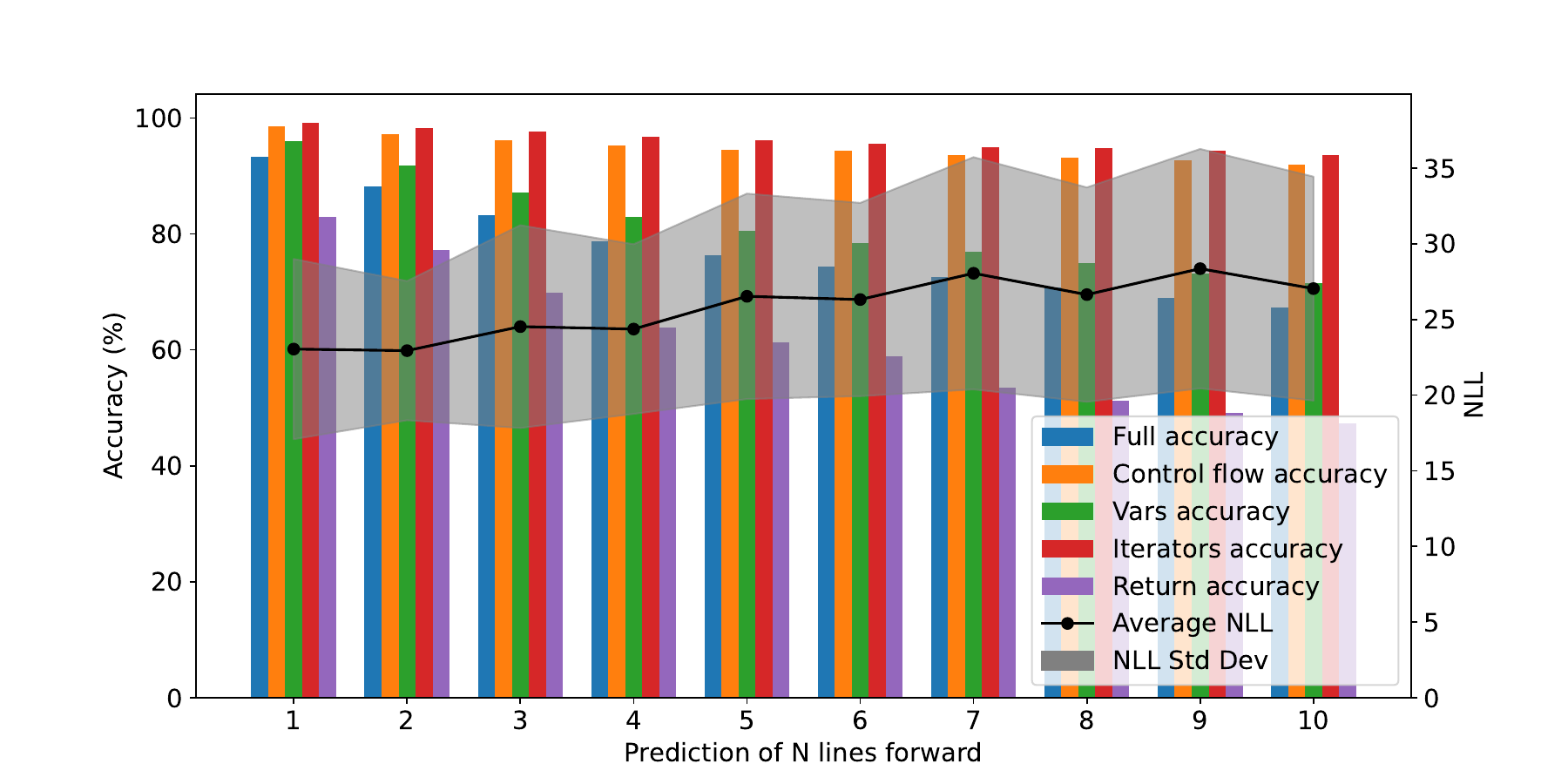}
    \caption{Plot showing \textit{individual} state prediction accuracy (e.g., for \textit{Return}, specifically for this plot and unlike in the rest of the article, we mean return statement accuracy, not full execution accuracy) when increasing N lines into the future, compared to the predictions Negative Log-Likelihood. Accuracy (bars) gets lower as the number of steps into the future increases, and confidence decreases as well (i.e., NLL increases).
    }
    \label{fig:variable_ns}

\end{wrapfigure}

\subsection{Individual state predictions on CruxEval}

Table \ref{tab:intrinsic_crux_e} shows the evaluations of individual states (not aggregated into full executions) for prompted out-of-the-box Llama 3.1 8B \citep{dubey2024llama3herdmodels}  and fine-tunings with traces on top of Llama 3.1 8B. In this section, for Line-n models, we report the average over n. 
\paragraph{Prompted (untrained) models} We find that general-purpose LLMs already exhibit non-trivial capabilities to predict execution steps out of the box. For example, Llama 3.1 8B prompted (i.e., not fine-tuned) to predict the state after executing the next line (Line-1 in Table \ref{tab:intrinsic_crux_e}) correctly guesses the answer 53.8\% of the times, implying a decent understanding of control flow. Control flow is considerably easier when prompted to predict the next bytecode state (Instruction-1 in Table \ref{tab:intrinsic_crux_e}), with a 74.1\%, since non-jump (and non-function calling) bytecode instructions have a linear flow. These control flow capabilities drop to 16.8\% when evaluated on \{1..10\} (averaged) lines into the future (Line-n in Table \ref{tab:intrinsic_crux_e}, i.e. asking the model to predict the state after N lines). Similar results, albeit slightly worse, are obtained for the iterator states. When looking at variable value predictions (Vars column), however, performance drops substantially for the line-level scratchpad. This struggle compared to other mainstream coding benchmarks hints at a lack of execution traces data in general-purpose LLMs. Notably, variable prediction for prompted Llama greatly improves for the Instruction-level variant (80.4\%). However, this is due to the heavy lifting being carried out on the stack, as variable states typically just consist of reading values from the stack into the variables. The stack-level accuracy is indeed low (5.8\%).

\textbf{Execution-tuned Models} Unsurprisingly, models trained on execution traces excel at control flow prediction. In particular, dynamic scratchpad models obtain almost perfect accuracy on control flow prediction, both for line (99.5\%) and instruction levels (99.9\%). The control flow accuracy only drops to 95.1\% for Line-n models, suggesting that the model is indeed capable of internally modeling the flow of future states. In comparison, scratchpad obtains a similarly high 91.0\%. Looking at iterator state prediction, both line and instruction-level dynamic scratchpads obtain almost perfect accuracy as well. Remarkably, the instruction-level model obtains an also near-perfect accuracy for the stack (98.8\%), in contrast to the prompted model.

\textbf{Skipping steps} Figure \ref{fig:variable_ns} shows the state prediction accuracy when increasing the number of states into the future and the corresponding negative log-likelihood (NLL) as a measure of the confidence of the prediction, for the Line-N dynamic scratchpad. Unsuprisingly, accuracy lowers as N increases. Interestingly, the corresponding NLL increases, showing calibration of the model confidence. Remarkably, however, we do not observe sharp drops in performance when looking at N steps into the future. Actually, our model can effectively learn to predict multiple steps into the future. Control flow and iterator states are relatively feasible to predict when jumping multiple steps, but variables and return values get increasingly complicated. In the Appendix, we provide similar results for Instruction-N.

In summary, while the out-of-the-box, prompted Llama shows non-trivial trace modeling capabilities (10.6\% full state accuracy with the Line-1 approach), models trained on traces greatly improve upon it. Interestingly, we observe that the line-based dynamic scratchpad outperforms (96.3\% full accuracy) its scratchpad counterpart (86.4\% full state accuracy), and that the instruction-level obtains the highest full state accuracy, 98.8\%. We also observe that the task of learning the state of N steps into the future is feasible to learn effectively, and that NLL has potential as a measure of model confidence in this setting. However, it remains to be seen how these  individual trace results will aggregate into function output predictions, which we study in the following sections.

\subsection{CruxEval output prediction}

\begin{table*}[thbp!]
\centering
\caption{\label{tab:crux_o}
CruxEval output prediction results, allowing for multi-step predictions  for the variants trained with execution traces.  *Global search using \citet{dijkstra1959note} the algorithm. Not directly comparable due to having access to the ground truth for checking correctness of paths.}
\scalebox{0.87}{
\begin{tabular}{llll}
\hline
\textsc{Representation} & \textsc{Outcome Accuracy} & \textsc{Process Accuracy} & \textsc{avg Steps needed} \\
\hline
Output FT & 49.3\% & - & 1 (direct)\\

\hline

Scratchpad F.T. & 78.7\% & 75.5\% & 10.8 lines \\
Compact Scratchpad F.T. & \textbf{79.7\%} & \textbf{76.6\%} & 11.8 lines \\

 Line-1 FT & 73.3\% & 73.3\% & 8.3 lines\\
Line-n FT & 60.8\% & 60.8\% & \textbf{2.9} lines \\
\hspace{3mm} + search* & 70.3\% & 70.3\% & \textbf{1.8} lines \\

Llama 3.1 8B + Instruction-1 F.T & 73.5\% & 73.5\% & 38.8 instructions  \\
\hspace{3mm} + search*  & 74.1\% & 74.1\% & 38.6 instructions \\
Llama 3.1 8B + Instruction-n F.T.  & 62.5\% & 62.5\% & \textbf{22.4} instructions  \\
\hspace{3mm} + search* & 73.5\% & 73.5\% & \textbf{4.8} instructions \\

\hline

Prompted Llama 3.1 8B** & 37.8\% & - & -\\ 
Prompted GPT-4 & 82\% & - & -\\

\hline
\end{tabular}
}

\end{table*}

We have seen the accuracy of the models when evaluated on individual state predictions. Here, we aggregate the results to evaluate output prediction on CruxEval in Table \ref{tab:crux_o}. For dynamic scratchpad models with more than one possible path (e.g., Line-n), we evaluate taking the \texttt{argmin(NLL)} one. Interestingly, the most confident prediction is not always the next immediate step, which is why predicting multiple steps ahead can lead to fewer overall steps.
We also show results with global search using \citet{dijkstra1959note}'s algorithm to obtain the shortest path using model predictions from the input of the function to the output,  which is not directly comparable to the other results due to having access to the ground truth for checking the correctness of paths. However, we provide it as an upper bound of what could be achieved with the predictions of the model. As a reference, we also  provide the top results in the CruxEval leaderboard, prompted GPT-4 with 82\% accuracy.\footnote{Pass-at-1, \texttt{gpt-4-turbo-2024-04-09+cot (n=3)}as of October 2024}

\textbf{Direct prediction} Out of the box, Llama 3.1 8B obtains an output prediction accuracy of 37.8\%. This accuracy can be improved to 49.3\% by fine tuning on direct output prediction. However, even with the relatively short executions found in CruxEval, more than half of the functions are out of the reach of the direct output prediction model.

\textbf{Results when using traces} Consistently with \citep{scratchpad}, all models trained on execution traces outperform by a great margin the direct output prediction fine-tuning. While we  generally obtain high accuracies (up to almost 80\%), note that the accuracy here, in Table \ref{tab:crux_o}, is substantially lower than in Table \ref{tab:intrinsic_crux_e}. The reason why this happens is that when aggregating individual trace predictions, a single step error (out of, e.g., 20 steps) can lead to a wrong result.   

\textbf{Comparison between models using traces} The compact scratchpad strategy slightly outperforms the full scratchpad one, and in turn these two outperform the dynamic scratchpad approaches. The executions in CruxEval are not long enough to show the advantages of dynamic scratchpads.  

\textbf{Indexing and string manipulation failure modes} In CruxEval, arithmetic operations (a classic failure mode of LLMs) were intentionally left out of the benchmark, to focus on program understanding itself. However, we noticed two interesting failure modes. After a qualitative error analysis, we found that the majority of the errors of the models on CruxEval belong to either one of two categories. The first one is string indexing. Indexing arbitrary strings is hard due to tokenization artifacts, since literals are tokenized inconsistently, and unlike arrays their elements aren't separated by punctuation.
However, it can be particularly hard for the dynamic scratchpad models (and this mainly explains the $\sim$ 5\% gap in output prediction accuracy between the line-level scratchpad and its dynamic scratchpad counterpart), because for each iteration the model needs to count from scratch to which characters the code is referring. Instead, the scratchpad model relies on the previous iteration as a hint to guess what character will come next. 
The second failure mode we saw consists of basic string manipulation. For example, models sometimes fail to predict the return value of Python's built-in \texttt{[string].istitle()} method, an issue that we also observed in the base model. CruxEval's string values might be out-of-distribution for the model.

\textbf{Skipping steps} Looking at the results of Line-n and Instruction-n, we observe that just by selecting the $n$ where the model is the most confident (based on NLL), we are able to obtain reasonable  accuracies (significantly better than direct prediction, albeit worse than Line-1 and Instruction-1) and considerably lesser number of steps needed. For example, Line-n on average needs only a 35\% of the steps of Line-1 to correctly predict a function. This has the remarkable implication that the ordering of model confidence for n=\{1..10\} does not always correspond to the number of steps into the future. That is, with significant frequency, $n=1$ is not always the prediction in which the model is the most confident.

\subsection{MBPP}

Next, we evaluate on the Python synthesis dataset MBPP \citep{mbpp} with the goal of observing results in longer executions and different domain as CruxEval. 

Particularly, we select functions in the MBPP test set with nested loops (as a  proxy of computational complexity and execution length), leaving us with slightly fewer than 100 functions.\footnote{Unfortunately, for MBPP we discovered an issue with traced iterators in nested \texttt{for} loops. Thus, specifically for MBPP we applied an AST transformation to rewrite nested \texttt{for} loops to \texttt{while} loops. This issue had virtually no effect in CruxEval, due to the lack of executed nested for loops.} 

Similarly to the case of CruxEval, Table \ref{tab:mbpp} shows the results on output prediction for this MBPP subset. The big picture of the results is similar to the case of CruxEval, but with some crucial differences. First, if we look at the average steps needed for correct predictions, we see that here the functions are indeed considerably longer than in CruxEval (in the order of 7x more executed lines). However, the lengths are still not astronomical. 
Relatively to the CruxEval results, here the instruction-based models perform considerably better, which we attribute to the fact that in MBPP there are computations that can be broken down into multiple instructions. Instead, in CruxEval, since most errors consist of indexing or guessing the outputs of string built-in methods, further zooming in doesn't help, as the (indexing or calling a string built-in written in C) can't be further divided.
Since in this benchmark some functions present auxiliary functions, we introduce a variant of the compact scratchpad in this the model is able to step in other called functions, yielding  an improvement of 3 points with respect to compact scratchpad (80.6\% vs. 77.4\%).

\begin{table*}[thbp!]
\centering
\caption{\label{tab:mbpp}
Evaluation on MBPP test set on functions with nested loops  }
\scalebox{0.87}{
\begin{tabular}{llll}
\hline
\textsc{Representation} & \textsc{Outcome Accuracy} & \textsc{Process Accuracy} & \textsc{avg Steps needed} \\
\hline
Output  F.T. & 47.3\%  & - & 1 (direct)\\

\hline

Scratchpad  F.T. & 64.5\% & 64.5\%  & 58.2  lines  \\
Compact Scratchpad F.T. & 77.4\%& 76.3\%  & 73.9  lines  \\
Compact Scratchpad +step-in F.T. & \textbf{80.6\%}& 74.2\%  & 73.9  lines  \\

Line-1 F.T. & 73.1\%  &73.1\%  & 73.8 lines\\
Line-n F.T. & 43\% & 43\% & 15.4  lines \\
\hspace{3mm} + search* & 59.1\%  & 59.1\%  & 7.8  lines \\

Instruction-1 F.T. & 78.5\% & 78.5\%  & 351.3 instructions  \\
\hspace{3mm} + search*  & \textbf{80.6\%} & 80.6\%  & 354.6  instructions \\
Instruction-n F.T. &65.6\% &65.6\% & 139.8 instructions  \\
\hspace{3mm} + search* & 88.2\%  & 88.2\% & 35  instructions \\

\end{tabular}
}

\end{table*}

\subsection{Long executions}

We observe that existing benchmarks for output prediction don't feature long executions. This is especially true for the standard one, CruxEval, but even when targeting functions with nested loops on MBPP, we rarely get to executions with more than 100 executed lines. In this section, we study well-known algorithmic tasks where we can obtain arbitrarily long executions: 
\begin{enumerate*}
    \item Collatz conjecture: a function returning the number of iterations required to reach 1 following the Collatz conjecture sequence, given a starting natural number.
    \item Binary counter: A 4-bit binary counter.
    \item Iterative Fibonacci: An iterative implementation of Fibonacci.
\end{enumerate*}

For selecting the inputs, we generate 4 random numbers (as the small inputs) between 1 and 20, and 5 between 20 and 4000 (as the larger inputs), and evaluate on all of them across the 3 functions. For Fibonacci, we restrict the evaluation on the smaller 5 numbers. For all functions in this section, we replace the function name by \texttt{f}, to give less hints to the model based on potential memorizations of well-known functions during pretraining. Table \ref{tab:summary} shows the summarized results on these tasks (see Appendix \ref{app:long_res_fine} for the fine-grained results).

\textbf{Collatz} The direct output prediction model is able to correctly predict the number of Collatz iterations for the 4 smaller numbers (up to $n=18$), and breaks for larger inputs. Curiously, the scratchpad model is not able to improve on the results of the direct output prediction model, and gets the same accuracy for a considerably increased number of intermediate steps (35 on average, corresponding to the number of executed lines). The compact scratchpad unlocks a larger input, $n=103$, for which is able to do 353 correct intermediate predictions, up to the (correct return value). 
The dynamic scratchpad models shine in this setting. Line-1 is able to correctly predict all studied inputs but 2620 (the next to largest one). For the largest input, $n=3038$, Line-1 needs to chain 619 correct predictions in a row.  Notably, Line-n is able to achieve the same accuracy but with only 39\% of the steps required by Line-1. Optimally, if we had access to an oracle that told us which of the paths was correct, Dijsktra would have yielded a perfect accuracy with only 32.3 steps required on average (compared to the average of 271 for Line-1).
\begin{wraptable}{r}{0.65\textwidth} 
\centering
\caption{\label{tab:summary} Long execution results: accuracy (avg. steps needed).}
\scalebox{0.85}{
\begin{tabular}{l|c|c|c}
\hline
\textsc{Representation} & \textsc{Collatz} & \textsc{Binary Counter} & \textsc{Fibonacci} \\
\hline
Output & 4/9 (1) & 1/9 (1) & 4/5 (1) \\
Scratchpad & 4/9 (35) & 4/9 (45.5) & 4/5 (37) \\
Compact Scratchpad & 5/9 (98.6) & 5/9 (116.2) & 5/5 (140.5) \\
Line-1 & 8/9 (271) & \textbf{8/9 (2441)} & 5/5 (140.5) \\
Line-n & \textbf{8/9 (106.1)} & 1/9 (6) & \textbf{5/5 (46.8)} \\
Line-n + Dijkstra & \textbf{9/9 (32.3)} & \textbf{9/9 (410.7)} & \textbf{5/5 (11.8)} \\
\hline
\end{tabular}
}
\end{wraptable}
\textbf{Binary counter} In the case of the 4-bit binary counter, curiously, the direct output prediction model is only able to correctly predict the output for the third smallest input ($n=8$). In this case, scratchpad does significantly improve results with respect to the direct output prediction model, correctly guessing the outputs for the 4 smaller inputs (up to $n=18$). However, the compact scratchpad is still better, unlocking the correct prediction for a bigger input, $n=103$. Curiously, Line-1 gets the same accuracy as with Collatz (all correct but the next to largest input), but with one crucial difference. Here, the executions are even longer. For correctly predicting the output for $n=3038$, Line-1 has to chain as many as 14,055 correct predictions in a row. Here, Line-n is not able to reliably select across paths based on its confidence (NLL). With Dijkstra on Line-n predictions, it would have obtained perfect accuracy with only 17\% of the number of steps needed by Line-1 on average.

\textbf{Fibonacci} Again, scratchpad obtains the exact same accuracy as direct prediction. The rest of the models are able to predict all inputs up to $n=103$. While both Compact scratchpad and Line-1 require 414 steps for $n=103$, Line-n is able to decrease this number to only 160 steps. Optimally, Dijkstra would have obtained 42.

\section{Downstream effects}
\label{sec:extrinsic}

So far we have shown that \name{} leads to improved output prediction capabilities. 
Here, we study its effects in a code supervised fine-tuning (SFT)  setting. 
We take the base Llama 3.1 8B \citep{dubey2024llama3herdmodels}, and evaluate the downstream performance with and without different versions of \name{} in the data mix. The base mix is a small code-only SFT dataset of samples similar to the ones in \citet{roziere2024codellamaopenfoundation}. We train for 7.5k steps with a global batch size of 1024 sequences of up to 8192 tokens.
We evaluate code generation on HumanEval \citep{chen2021codex} and MBPP \citep{mbpp}, without including the traces in inference time. We also evaluate a step-by-step reasoning task, GSM8k\citep{cobbe2021gsm8k}, to study 
 potential improved multi-step reasoning in other domains. 

Table \ref{tab:downstream} shows downstream evaluation results with and without \name{} in the SFT mix. Fine-tuning on direct I/O prediction improves Crux-I and Crux-O but not coding benchmarks. The best-performing trace variant, with 10\% Compact Scratchpad, brings slight gains on HumanEval, MBPP, and 1.2 points on GSM8K. Curiously, forward execution fine-tuning worsens Crux-I, and vice versa, suggesting weaker-than-expected ties  between forward and backward prediction. These results indicate that merging \name{} with SFT data offers little coding improvement. We hypothesize increased gains in evaluations  related to program state, such as  test generation or debugger-assisted tasks.

\begin{table}[thbp!]
  \center
  \caption{ Downstream evaluations on HumanEval, MBPP and GSM8K.
  \label{tab:downstream}}
  \scalebox{0.74}{
   \setlength{\tabcolsep}{3pt}
  \begin{tabular}{l|cc|cc|ccc|ccc|c}
  \toprule
  Model & \multicolumn{2}{c}{Crux-I} & \multicolumn{2}{c}{Crux-O} & \multicolumn{3}{c}{HumanEval} & \multicolumn{3}{c}{MBPP} & GSM8K \\
  & pass@1 & pass@5 & pass@1 & pass@5   & pass@1 & pass@10 & pass@100 & pass@1 & pass@10 & pass@100 & 0-shot\\
  \midrule 
 SFT mix  & 42.9 & 56.5 & 36.4 & 52 & 56.7 & 79.1 &  \textbf{91.2} & 52.2 & 69.4 & 80.7 & 66.3\\ 
  \midrule
   \hspace{3mm} + input FT (10\%)   & \textbf{43.8} & \textbf{57.9} & 34.8  & 46.8  & 51.8 & 77.6  &90.7  & 52.3  & 68.2 & 79.4 & 66\\ 
    \hspace{3mm} + output FT (10\%)    & 41.1 & 56.2 & \textbf{41.8} & \textbf{52.5}  & 56.7  & 79.2  & 91 & 23.8  & 65.2 & 79.1 & 65.1 \\ 
     \midrule
    \hspace{3mm} + C. Scratch  (10\%)  & 41.8 & 56 & 38.8 & 49.8 & \textbf{58.5}  & \textbf{79.9}  & 89.9  & \textbf{53}  & \textbf{69.6} & \textbf{82.5} & \textbf{67.5} \\ 
        \hspace{3mm} + C. Scratch (5\%)  & 42.9 & 58.3 & 38 & 50.5 & 57.9 & 78.9  & 89.3 & 52.6  & 69.3 & 81 & 66.3 \\
        \hspace{3mm} + Line-1 (10\%)    & 39.8 &  56.5& 38 & 50.5  &  53.7 &  78.7  &  89.2  & \textbf{53}  & 69.2 & 80.2 & 65 \\
    \hspace{3mm} + Line-n  (10\%)   & 39.4 & 56.3& 38.8 & 49.1  &  56.1&  78.3 &  88.7  & 51.4  & 68.5 & 80.8 & 66.2 \\

  \bottomrule
  \end{tabular}
  }
  
\end{table}

\section{Related work}
\label{sec:related}

Learning to execute programs as  a benchmarking task for code reasoning capabilities has been long studied in the machine learning community \citep{learning_to_execute2014}, sometimes with niche architectures
\citep{GravesWD14, GauntBKT16, learning_to_execute2020}, typically on toy or restricted programs. \citet{bieber2022staticpredictionruntimeerrors} proposed learning to predict runtime errors as a practical application of neural program evaluation. More recently, \citet{gu2024cruxeval} introduced  a program output (and input) benchmark for LLMs to measure code understanding capabilities, which we used for evaluation in this work.
Most closely to ours, \citet{scratchpad} propose the use of \textit{scratchpads} to let LLMs write down the results of intermediate computations rather than directly aiming at predicting the final output. With Python output prediction being one of their use cases, they represent traces of intermediate states as JSON dictionaries. 
\citet{ni2024nextteachinglargelanguage} introduce \textit{Naturalized} EXecution Tuning (NExT) and propose the compact representation of Python traces that we followed. Unlike \citet{scratchpad} and this work, NeXT simplifies loops and uses traces in the \textit{input}, improving program repair.  \citet{ding2024semcodertrainingcodelanguage} propose natural language explanations based on executions, leading to further improvements. 
Finally, recent work uses execution \textit{feedback}, rather than traces, in SFT or reinforcement learning settings \citep{dong2024selfplayexecutionfeedbackimproving,gehring2024rlefgroundingcodellms}.

\section{Conclusion}
\label{sec:conclusion}

In this work, we conducted a large-scale study on traces modeling building upon \citet{scratchpad} and \citet{ni2024nextteachinglargelanguage}. Reflecting back on our questions, (1) we can scale trace modeling up with \name{}, by tracing executions on automatically generated inputs and thus generating large training datasets, which generalizes to output prediction benchmarks. (2) A more fine-grained granularity can't help when the core issue can't be further broken down (indexing on CruxEval) but shows promise otherwise. Regarding scratchpad strategies (3) and execution lengths (4), our newly introduced dynamic scratchpad excels at very long executions, while compact scratchpad generally outperforms the original scratchpad. We saw no conclusive improvements on downstream coding benchmarks (5), where program state understanding might not be critical. 
As future work, we suggest extending our work to other languages such as C, pointer ids to understand phenomena such as aliasing, and closures. We are also keen on more challenging datasets with exceptions and dynamic trace granularities.

%

\clearpage

\bibliography{main}

\begin{thebibliography}{22}
\providecommand{\natexlab}[1]{#1}
\providecommand{\url}[1]{\texttt{#1}}
\expandafter\ifx\csname urlstyle\endcsname\relax
  \providecommand{\doi}[1]{doi: #1}\else
  \providecommand{\doi}{doi: \begingroup \urlstyle{rm}\Url}\fi

\bibitem[Aryabumi et~al.(2024)Aryabumi, Su, Ma, Morisot, Zhang, Locatelli, Fadaee, Üstün, and Hooker]{aryabumi2024codecodeexploringimpact}
Viraat Aryabumi, Yixuan Su, Raymond Ma, Adrien Morisot, Ivan Zhang, Acyr Locatelli, Marzieh Fadaee, Ahmet Üstün, and Sara Hooker.
\newblock To code, or not to code? exploring impact of code in pre-training, 2024.
\newblock URL \url{https://arxiv.org/abs/2408.10914}.

\bibitem[Austin et~al.(2021{\natexlab{a}})Austin, Odena, Nye, Bosma, Michalewski, Dohan, Jiang, Cai, Terry, Le, and Sutton]{Austin2021}
Jacob Austin, Augustus Odena, Maxwell~I. Nye, Maarten Bosma, Henryk Michalewski, David Dohan, Ellen Jiang, Carrie~J. Cai, Michael Terry, Quoc~V. Le, and Charles Sutton.
\newblock Program synthesis with large language models.
\newblock \emph{CoRR}, abs/2108.07732, 2021{\natexlab{a}}.
\newblock URL \url{https://arxiv.org/abs/2108.07732}.

\bibitem[Austin et~al.(2021{\natexlab{b}})Austin, Odena, Nye, Bosma, Michalewski, Dohan, Jiang, Cai, Terry, Le, and Sutton]{mbpp}
Jacob Austin, Augustus Odena, Maxwell~I. Nye, Maarten Bosma, Henryk Michalewski, David Dohan, Ellen Jiang, Carrie~J. Cai, Michael Terry, Quoc~V. Le, and Charles Sutton.
\newblock Program synthesis with large language models.
\newblock \emph{CoRR}, abs/2108.07732, 2021{\natexlab{b}}.
\newblock URL \url{https://arxiv.org/abs/2108.07732}.

\bibitem[Bieber et~al.(2020)Bieber, Sutton, Larochelle, and Tarlow]{learning_to_execute2020}
David Bieber, Charles Sutton, Hugo Larochelle, and Daniel Tarlow.
\newblock Learning to execute programs with instruction pointer attention graph neural networks.
\newblock \emph{CoRR}, abs/2010.12621, 2020.
\newblock URL \url{https://arxiv.org/abs/2010.12621}.

\bibitem[Bieber et~al.(2022)Bieber, Goel, Zheng, Larochelle, and Tarlow]{bieber2022staticpredictionruntimeerrors}
David Bieber, Rishab Goel, Daniel Zheng, Hugo Larochelle, and Daniel Tarlow.
\newblock Static prediction of runtime errors by learning to execute programs with external resource descriptions, 2022.
\newblock URL \url{https://arxiv.org/abs/2203.03771}.

\bibitem[Brown et~al.(2020)Brown, Mann, Ryder, Subbiah, Kaplan, Dhariwal, Neelakantan, Shyam, Sastry, Askell, Agarwal, Herbert-Voss, Krueger, Henighan, Child, Ramesh, Ziegler, Wu, Winter, Hesse, Chen, Sigler, Litwin, Gray, Chess, Clark, Berner, McCandlish, Radford, Sutskever, and Amodei]{NEURIPS2020_1457c0d6}
Tom Brown, Benjamin Mann, Nick Ryder, Melanie Subbiah, Jared~D Kaplan, Prafulla Dhariwal, Arvind Neelakantan, Pranav Shyam, Girish Sastry, Amanda Askell, Sandhini Agarwal, Ariel Herbert-Voss, Gretchen Krueger, Tom Henighan, Rewon Child, Aditya Ramesh, Daniel Ziegler, Jeffrey Wu, Clemens Winter, Chris Hesse, Mark Chen, Eric Sigler, Mateusz Litwin, Scott Gray, Benjamin Chess, Jack Clark, Christopher Berner, Sam McCandlish, Alec Radford, Ilya Sutskever, and Dario Amodei.
\newblock Language models are few-shot learners.
\newblock In H.~Larochelle, M.~Ranzato, R.~Hadsell, M.F. Balcan, and H.~Lin (eds.), \emph{Advances in Neural Information Processing Systems}, volume~33, pp.\  1877--1901. Curran Associates, Inc., 2020.
\newblock URL \url{https://proceedings.neurips.cc/paper_files/paper/2020/file/1457c0d6bfcb4967418bfb8ac142f64a-Paper.pdf}.

\bibitem[Chen et~al.(2021)Chen, Tworek, Jun, Yuan, de~Oliveira~Pinto, Kaplan, Edwards, Burda, Joseph, Brockman, Ray, Puri, Krueger, Petrov, Khlaaf, Sastry, Mishkin, Chan, Gray, Ryder, Pavlov, Power, Kaiser, Bavarian, Winter, Tillet, Such, Cummings, Plappert, Chantzis, Barnes, Herbert-Voss, Guss, Nichol, Paino, Tezak, Tang, Babuschkin, Balaji, Jain, Saunders, Hesse, Carr, Leike, Achiam, Misra, Morikawa, Radford, Knight, Brundage, Murati, Mayer, Welinder, McGrew, Amodei, McCandlish, Sutskever, and Zaremba]{chen2021codex}
Mark Chen, Jerry Tworek, Heewoo Jun, Qiming Yuan, Henrique~Ponde de~Oliveira~Pinto, Jared Kaplan, Harri Edwards, Yuri Burda, Nicholas Joseph, Greg Brockman, Alex Ray, Raul Puri, Gretchen Krueger, Michael Petrov, Heidy Khlaaf, Girish Sastry, Pamela Mishkin, Brooke Chan, Scott Gray, Nick Ryder, Mikhail Pavlov, Alethea Power, Lukasz Kaiser, Mohammad Bavarian, Clemens Winter, Philippe Tillet, Felipe~Petroski Such, Dave Cummings, Matthias Plappert, Fotios Chantzis, Elizabeth Barnes, Ariel Herbert-Voss, William~Hebgen Guss, Alex Nichol, Alex Paino, Nikolas Tezak, Jie Tang, Igor Babuschkin, Suchir Balaji, Shantanu Jain, William Saunders, Christopher Hesse, Andrew~N. Carr, Jan Leike, Josh Achiam, Vedant Misra, Evan Morikawa, Alec Radford, Matthew Knight, Miles Brundage, Mira Murati, Katie Mayer, Peter Welinder, Bob McGrew, Dario Amodei, Sam McCandlish, Ilya Sutskever, and Wojciech Zaremba.
\newblock Evaluating large language models trained on code.
\newblock 2021.

\bibitem[Cobbe et~al.(2021)Cobbe, Kosaraju, Bavarian, Chen, Jun, Kaiser, Plappert, Tworek, Hilton, Nakano, Hesse, and Schulman]{cobbe2021gsm8k}
Karl Cobbe, Vineet Kosaraju, Mohammad Bavarian, Mark Chen, Heewoo Jun, Lukasz Kaiser, Matthias Plappert, Jerry Tworek, Jacob Hilton, Reiichiro Nakano, Christopher Hesse, and John Schulman.
\newblock Training verifiers to solve math word problems.
\newblock \emph{arXiv preprint arXiv:2110.14168}, 2021.

\bibitem[Dijkstra(1959)]{dijkstra1959note}
Edsger~W Dijkstra.
\newblock A note on two problems in connexion with graphs.
\newblock \emph{Numerische mathematik}, 1\penalty0 (1):\penalty0 269--271, 1959.

\bibitem[Ding et~al.(2024)Ding, Peng, Min, Kaiser, Yang, and Ray]{ding2024semcodertrainingcodelanguage}
Yangruibo Ding, Jinjun Peng, Marcus~J. Min, Gail Kaiser, Junfeng Yang, and Baishakhi Ray.
\newblock Semcoder: Training code language models with comprehensive semantics, 2024.
\newblock URL \url{https://arxiv.org/abs/2406.01006}.

\bibitem[Dong et~al.(2024)Dong, Lu, Li, Xia, Yu, Zhou, and Zhou]{dong2024selfplayexecutionfeedbackimproving}
Guanting Dong, Keming Lu, Chengpeng Li, Tingyu Xia, Bowen Yu, Chang Zhou, and Jingren Zhou.
\newblock Self-play with execution feedback: Improving instruction-following capabilities of large language models, 2024.
\newblock URL \url{https://arxiv.org/abs/2406.13542}.

\bibitem[Dubey et~al.(2024)Dubey, Jauhri, Pandey, Kadian, Al-Dahle, Letman, Mathur, Schelten, Yang, Fan, Goyal, Hartshorn, Yang, Mitra, Sravankumar, Korenev, Hinsvark, Rao, Zhang, Rodriguez, Gregerson, Spataru, Roziere, Biron, Tang, Chern, Caucheteux, Nayak, Bi, Marra, McConnell, Keller, Touret, Wu, Wong, Ferrer, Nikolaidis, Allonsius, Song, Pintz, Livshits, Esiobu, Choudhary, Mahajan, Garcia-Olano, Perino, Hupkes, Lakomkin, AlBadawy, Lobanova, Dinan, Smith, Radenovic, Zhang, Synnaeve, Lee, Anderson, Nail, Mialon, Pang, Cucurell, Nguyen, Korevaar, Xu, Touvron, Zarov, Ibarra, Kloumann, Misra, Evtimov, Copet, Lee, Geffert, Vranes, Park, Mahadeokar, Shah, van~der Linde, Billock, Hong, Lee, Fu, Chi, Huang, Liu, Wang, Yu, Bitton, Spisak, Park, Rocca, Johnstun, Saxe, Jia, Alwala, Upasani, Plawiak, Li, Heafield, Stone, El-Arini, Iyer, Malik, Chiu, Bhalla, Rantala-Yeary, van~der Maaten, Chen, Tan, Jenkins, Martin, Madaan, Malo, Blecher, Landzaat, de~Oliveira, Muzzi, Pasupuleti, Singh, Paluri, Kardas, Oldham, Rita,
  Pavlova, Kambadur, Lewis, Si, Singh, Hassan, Goyal, Torabi, Bashlykov, Bogoychev, Chatterji, Duchenne, Çelebi, Alrassy, Zhang, Li, Vasic, Weng, Bhargava, Dubal, Krishnan, Koura, Xu, He, Dong, Srinivasan, Ganapathy, Calderer, Cabral, Stojnic, Raileanu, Girdhar, Patel, Sauvestre, Polidoro, Sumbaly, Taylor, Silva, Hou, Wang, Hosseini, Chennabasappa, Singh, Bell, Kim, Edunov, Nie, Narang, Raparthy, Shen, Wan, Bhosale, Zhang, Vandenhende, Batra, Whitman, Sootla, Collot, Gururangan, Borodinsky, Herman, Fowler, Sheasha, Georgiou, Scialom, Speckbacher, Mihaylov, Xiao, Karn, Goswami, Gupta, Ramanathan, Kerkez, Gonguet, Do, Vogeti, Petrovic, Chu, Xiong, Fu, Meers, Martinet, Wang, Tan, Xie, Jia, Wang, Goldschlag, Gaur, Babaei, Wen, Song, Zhang, Li, Mao, Coudert, Yan, Chen, Papakipos, Singh, Grattafiori, Jain, Kelsey, Shajnfeld, Gangidi, Victoria, Goldstand, Menon, Sharma, Boesenberg, Vaughan, Baevski, Feinstein, Kallet, Sangani, Yunus, Lupu, Alvarado, Caples, Gu, Ho, Poulton, Ryan, Ramchandani, Franco, Saraf,
  Chowdhury, Gabriel, Bharambe, Eisenman, Yazdan, James, Maurer, Leonhardi, Huang, Loyd, Paola, Paranjape, Liu, Wu, Ni, Hancock, Wasti, Spence, Stojkovic, Gamido, Montalvo, Parker, Burton, Mejia, Wang, Kim, Zhou, Hu, Chu, Cai, Tindal, Feichtenhofer, Civin, Beaty, Kreymer, Li, Wyatt, Adkins, Xu, Testuggine, David, Parikh, Liskovich, Foss, Wang, Le, Holland, Dowling, Jamil, Montgomery, Presani, Hahn, Wood, Brinkman, Arcaute, Dunbar, Smothers, Sun, Kreuk, Tian, Ozgenel, Caggioni, Guzmán, Kanayet, Seide, Florez, Schwarz, Badeer, Swee, Halpern, Thattai, Herman, Sizov, Guangyi, Zhang, Lakshminarayanan, Shojanazeri, Zou, Wang, Zha, Habeeb, Rudolph, Suk, Aspegren, Goldman, Damlaj, Molybog, Tufanov, Veliche, Gat, Weissman, Geboski, Kohli, Asher, Gaya, Marcus, Tang, Chan, Zhen, Reizenstein, Teboul, Zhong, Jin, Yang, Cummings, Carvill, Shepard, McPhie, Torres, Ginsburg, Wang, Wu, U, Saxena, Prasad, Khandelwal, Zand, Matosich, Veeraraghavan, Michelena, Li, Huang, Chawla, Lakhotia, Huang, Chen, Garg, A, Silva, Bell,
  Zhang, Guo, Yu, Moshkovich, Wehrstedt, Khabsa, Avalani, Bhatt, Tsimpoukelli, Mankus, Hasson, Lennie, Reso, Groshev, Naumov, Lathi, Keneally, Seltzer, Valko, Restrepo, Patel, Vyatskov, Samvelyan, Clark, Macey, Wang, Hermoso, Metanat, Rastegari, Bansal, Santhanam, Parks, White, Bawa, Singhal, Egebo, Usunier, Laptev, Dong, Zhang, Cheng, Chernoguz, Hart, Salpekar, Kalinli, Kent, Parekh, Saab, Balaji, Rittner, Bontrager, Roux, Dollar, Zvyagina, Ratanchandani, Yuvraj, Liang, Alao, Rodriguez, Ayub, Murthy, Nayani, Mitra, Li, Hogan, Battey, Wang, Maheswari, Howes, Rinott, Bondu, Datta, Chugh, Hunt, Dhillon, Sidorov, Pan, Verma, Yamamoto, Ramaswamy, Lindsay, Lindsay, Feng, Lin, Zha, Shankar, Zhang, Zhang, Wang, Agarwal, Sajuyigbe, Chintala, Max, Chen, Kehoe, Satterfield, Govindaprasad, Gupta, Cho, Virk, Subramanian, Choudhury, Goldman, Remez, Glaser, Best, Kohler, Robinson, Li, Zhang, Matthews, Chou, Shaked, Vontimitta, Ajayi, Montanez, Mohan, Kumar, Mangla, Albiero, Ionescu, Poenaru, Mihailescu, Ivanov, Li, Wang,
  Jiang, Bouaziz, Constable, Tang, Wang, Wu, Wang, Xia, Wu, Gao, Chen, Hu, Jia, Qi, Li, Zhang, Zhang, Adi, Nam, Yu, Wang, Hao, Qian, He, Rait, DeVito, Rosnbrick, Wen, Yang, and Zhao]{dubey2024llama3herdmodels}
Abhimanyu Dubey, Abhinav Jauhri, Abhinav Pandey, Abhishek Kadian, Ahmad Al-Dahle, Aiesha Letman, Akhil Mathur, Alan Schelten, Amy Yang, Angela Fan, Anirudh Goyal, Anthony Hartshorn, Aobo Yang, Archi Mitra, Archie Sravankumar, Artem Korenev, Arthur Hinsvark, Arun Rao, Aston Zhang, Aurelien Rodriguez, Austen Gregerson, Ava Spataru, Baptiste Roziere, Bethany Biron, Binh Tang, Bobbie Chern, Charlotte Caucheteux, Chaya Nayak, Chloe Bi, Chris Marra, Chris McConnell, Christian Keller, Christophe Touret, Chunyang Wu, Corinne Wong, Cristian~Canton Ferrer, Cyrus Nikolaidis, Damien Allonsius, Daniel Song, Danielle Pintz, Danny Livshits, David Esiobu, Dhruv Choudhary, Dhruv Mahajan, Diego Garcia-Olano, Diego Perino, Dieuwke Hupkes, Egor Lakomkin, Ehab AlBadawy, Elina Lobanova, Emily Dinan, Eric~Michael Smith, Filip Radenovic, Frank Zhang, Gabriel Synnaeve, Gabrielle Lee, Georgia~Lewis Anderson, Graeme Nail, Gregoire Mialon, Guan Pang, Guillem Cucurell, Hailey Nguyen, Hannah Korevaar, Hu~Xu, Hugo Touvron, Iliyan Zarov,
  Imanol~Arrieta Ibarra, Isabel Kloumann, Ishan Misra, Ivan Evtimov, Jade Copet, Jaewon Lee, Jan Geffert, Jana Vranes, Jason Park, Jay Mahadeokar, Jeet Shah, Jelmer van~der Linde, Jennifer Billock, Jenny Hong, Jenya Lee, Jeremy Fu, Jianfeng Chi, Jianyu Huang, Jiawen Liu, Jie Wang, Jiecao Yu, Joanna Bitton, Joe Spisak, Jongsoo Park, Joseph Rocca, Joshua Johnstun, Joshua Saxe, Junteng Jia, Kalyan~Vasuden Alwala, Kartikeya Upasani, Kate Plawiak, Ke~Li, Kenneth Heafield, Kevin Stone, Khalid El-Arini, Krithika Iyer, Kshitiz Malik, Kuenley Chiu, Kunal Bhalla, Lauren Rantala-Yeary, Laurens van~der Maaten, Lawrence Chen, Liang Tan, Liz Jenkins, Louis Martin, Lovish Madaan, Lubo Malo, Lukas Blecher, Lukas Landzaat, Luke de~Oliveira, Madeline Muzzi, Mahesh Pasupuleti, Mannat Singh, Manohar Paluri, Marcin Kardas, Mathew Oldham, Mathieu Rita, Maya Pavlova, Melanie Kambadur, Mike Lewis, Min Si, Mitesh~Kumar Singh, Mona Hassan, Naman Goyal, Narjes Torabi, Nikolay Bashlykov, Nikolay Bogoychev, Niladri Chatterji, Olivier
  Duchenne, Onur Çelebi, Patrick Alrassy, Pengchuan Zhang, Pengwei Li, Petar Vasic, Peter Weng, Prajjwal Bhargava, Pratik Dubal, Praveen Krishnan, Punit~Singh Koura, Puxin Xu, Qing He, Qingxiao Dong, Ragavan Srinivasan, Raj Ganapathy, Ramon Calderer, Ricardo~Silveira Cabral, Robert Stojnic, Roberta Raileanu, Rohit Girdhar, Rohit Patel, Romain Sauvestre, Ronnie Polidoro, Roshan Sumbaly, Ross Taylor, Ruan Silva, Rui Hou, Rui Wang, Saghar Hosseini, Sahana Chennabasappa, Sanjay Singh, Sean Bell, Seohyun~Sonia Kim, Sergey Edunov, Shaoliang Nie, Sharan Narang, Sharath Raparthy, Sheng Shen, Shengye Wan, Shruti Bhosale, Shun Zhang, Simon Vandenhende, Soumya Batra, Spencer Whitman, Sten Sootla, Stephane Collot, Suchin Gururangan, Sydney Borodinsky, Tamar Herman, Tara Fowler, Tarek Sheasha, Thomas Georgiou, Thomas Scialom, Tobias Speckbacher, Todor Mihaylov, Tong Xiao, Ujjwal Karn, Vedanuj Goswami, Vibhor Gupta, Vignesh Ramanathan, Viktor Kerkez, Vincent Gonguet, Virginie Do, Vish Vogeti, Vladan Petrovic, Weiwei Chu,
  Wenhan Xiong, Wenyin Fu, Whitney Meers, Xavier Martinet, Xiaodong Wang, Xiaoqing~Ellen Tan, Xinfeng Xie, Xuchao Jia, Xuewei Wang, Yaelle Goldschlag, Yashesh Gaur, Yasmine Babaei, Yi~Wen, Yiwen Song, Yuchen Zhang, Yue Li, Yuning Mao, Zacharie~Delpierre Coudert, Zheng Yan, Zhengxing Chen, Zoe Papakipos, Aaditya Singh, Aaron Grattafiori, Abha Jain, Adam Kelsey, Adam Shajnfeld, Adithya Gangidi, Adolfo Victoria, Ahuva Goldstand, Ajay Menon, Ajay Sharma, Alex Boesenberg, Alex Vaughan, Alexei Baevski, Allie Feinstein, Amanda Kallet, Amit Sangani, Anam Yunus, Andrei Lupu, Andres Alvarado, Andrew Caples, Andrew Gu, Andrew Ho, Andrew Poulton, Andrew Ryan, Ankit Ramchandani, Annie Franco, Aparajita Saraf, Arkabandhu Chowdhury, Ashley Gabriel, Ashwin Bharambe, Assaf Eisenman, Azadeh Yazdan, Beau James, Ben Maurer, Benjamin Leonhardi, Bernie Huang, Beth Loyd, Beto~De Paola, Bhargavi Paranjape, Bing Liu, Bo~Wu, Boyu Ni, Braden Hancock, Bram Wasti, Brandon Spence, Brani Stojkovic, Brian Gamido, Britt Montalvo, Carl
  Parker, Carly Burton, Catalina Mejia, Changhan Wang, Changkyu Kim, Chao Zhou, Chester Hu, Ching-Hsiang Chu, Chris Cai, Chris Tindal, Christoph Feichtenhofer, Damon Civin, Dana Beaty, Daniel Kreymer, Daniel Li, Danny Wyatt, David Adkins, David Xu, Davide Testuggine, Delia David, Devi Parikh, Diana Liskovich, Didem Foss, Dingkang Wang, Duc Le, Dustin Holland, Edward Dowling, Eissa Jamil, Elaine Montgomery, Eleonora Presani, Emily Hahn, Emily Wood, Erik Brinkman, Esteban Arcaute, Evan Dunbar, Evan Smothers, Fei Sun, Felix Kreuk, Feng Tian, Firat Ozgenel, Francesco Caggioni, Francisco Guzmán, Frank Kanayet, Frank Seide, Gabriela~Medina Florez, Gabriella Schwarz, Gada Badeer, Georgia Swee, Gil Halpern, Govind Thattai, Grant Herman, Grigory Sizov, Guangyi, Zhang, Guna Lakshminarayanan, Hamid Shojanazeri, Han Zou, Hannah Wang, Hanwen Zha, Haroun Habeeb, Harrison Rudolph, Helen Suk, Henry Aspegren, Hunter Goldman, Ibrahim Damlaj, Igor Molybog, Igor Tufanov, Irina-Elena Veliche, Itai Gat, Jake Weissman, James
  Geboski, James Kohli, Japhet Asher, Jean-Baptiste Gaya, Jeff Marcus, Jeff Tang, Jennifer Chan, Jenny Zhen, Jeremy Reizenstein, Jeremy Teboul, Jessica Zhong, Jian Jin, Jingyi Yang, Joe Cummings, Jon Carvill, Jon Shepard, Jonathan McPhie, Jonathan Torres, Josh Ginsburg, Junjie Wang, Kai Wu, Kam~Hou U, Karan Saxena, Karthik Prasad, Kartikay Khandelwal, Katayoun Zand, Kathy Matosich, Kaushik Veeraraghavan, Kelly Michelena, Keqian Li, Kun Huang, Kunal Chawla, Kushal Lakhotia, Kyle Huang, Lailin Chen, Lakshya Garg, Lavender A, Leandro Silva, Lee Bell, Lei Zhang, Liangpeng Guo, Licheng Yu, Liron Moshkovich, Luca Wehrstedt, Madian Khabsa, Manav Avalani, Manish Bhatt, Maria Tsimpoukelli, Martynas Mankus, Matan Hasson, Matthew Lennie, Matthias Reso, Maxim Groshev, Maxim Naumov, Maya Lathi, Meghan Keneally, Michael~L. Seltzer, Michal Valko, Michelle Restrepo, Mihir Patel, Mik Vyatskov, Mikayel Samvelyan, Mike Clark, Mike Macey, Mike Wang, Miquel~Jubert Hermoso, Mo~Metanat, Mohammad Rastegari, Munish Bansal, Nandhini
  Santhanam, Natascha Parks, Natasha White, Navyata Bawa, Nayan Singhal, Nick Egebo, Nicolas Usunier, Nikolay~Pavlovich Laptev, Ning Dong, Ning Zhang, Norman Cheng, Oleg Chernoguz, Olivia Hart, Omkar Salpekar, Ozlem Kalinli, Parkin Kent, Parth Parekh, Paul Saab, Pavan Balaji, Pedro Rittner, Philip Bontrager, Pierre Roux, Piotr Dollar, Polina Zvyagina, Prashant Ratanchandani, Pritish Yuvraj, Qian Liang, Rachad Alao, Rachel Rodriguez, Rafi Ayub, Raghotham Murthy, Raghu Nayani, Rahul Mitra, Raymond Li, Rebekkah Hogan, Robin Battey, Rocky Wang, Rohan Maheswari, Russ Howes, Ruty Rinott, Sai~Jayesh Bondu, Samyak Datta, Sara Chugh, Sara Hunt, Sargun Dhillon, Sasha Sidorov, Satadru Pan, Saurabh Verma, Seiji Yamamoto, Sharadh Ramaswamy, Shaun Lindsay, Shaun Lindsay, Sheng Feng, Shenghao Lin, Shengxin~Cindy Zha, Shiva Shankar, Shuqiang Zhang, Shuqiang Zhang, Sinong Wang, Sneha Agarwal, Soji Sajuyigbe, Soumith Chintala, Stephanie Max, Stephen Chen, Steve Kehoe, Steve Satterfield, Sudarshan Govindaprasad, Sumit Gupta,
  Sungmin Cho, Sunny Virk, Suraj Subramanian, Sy~Choudhury, Sydney Goldman, Tal Remez, Tamar Glaser, Tamara Best, Thilo Kohler, Thomas Robinson, Tianhe Li, Tianjun Zhang, Tim Matthews, Timothy Chou, Tzook Shaked, Varun Vontimitta, Victoria Ajayi, Victoria Montanez, Vijai Mohan, Vinay~Satish Kumar, Vishal Mangla, Vítor Albiero, Vlad Ionescu, Vlad Poenaru, Vlad~Tiberiu Mihailescu, Vladimir Ivanov, Wei Li, Wenchen Wang, Wenwen Jiang, Wes Bouaziz, Will Constable, Xiaocheng Tang, Xiaofang Wang, Xiaojian Wu, Xiaolan Wang, Xide Xia, Xilun Wu, Xinbo Gao, Yanjun Chen, Ye~Hu, Ye~Jia, Ye~Qi, Yenda Li, Yilin Zhang, Ying Zhang, Yossi Adi, Youngjin Nam, Yu, Wang, Yuchen Hao, Yundi Qian, Yuzi He, Zach Rait, Zachary DeVito, Zef Rosnbrick, Zhaoduo Wen, Zhenyu Yang, and Zhiwei Zhao.
\newblock The llama 3 herd of models, 2024.
\newblock URL \url{https://arxiv.org/abs/2407.21783}.

\bibitem[Gaunt et~al.(2016)Gaunt, Brockschmidt, Kushman, and Tarlow]{GauntBKT16}
Alexander~L. Gaunt, Marc Brockschmidt, Nate Kushman, and Daniel Tarlow.
\newblock Lifelong perceptual programming by example.
\newblock \emph{CoRR}, abs/1611.02109, 2016.
\newblock URL \url{http://arxiv.org/abs/1611.02109}.

\bibitem[Gehring et~al.(2024)Gehring, Zheng, Copet, Mella, Cohen, and Synnaeve]{gehring2024rlefgroundingcodellms}
Jonas Gehring, Kunhao Zheng, Jade Copet, Vegard Mella, Taco Cohen, and Gabriel Synnaeve.
\newblock Rlef: Grounding code llms in execution feedback with reinforcement learning, 2024.
\newblock URL \url{https://arxiv.org/abs/2410.02089}.

\bibitem[Graves et~al.(2014)Graves, Wayne, and Danihelka]{GravesWD14}
Alex Graves, Greg Wayne, and Ivo Danihelka.
\newblock Neural turing machines.
\newblock \emph{CoRR}, abs/1410.5401, 2014.
\newblock URL \url{http://arxiv.org/abs/1410.5401}.

\bibitem[Gu et~al.(2024)Gu, Rozière, Leather, Solar-Lezama, Synnaeve, and Wang]{gu2024cruxeval}
Alex Gu, Baptiste Rozière, Hugh Leather, Armando Solar-Lezama, Gabriel Synnaeve, and Sida~I. Wang.
\newblock Cruxeval: A benchmark for code reasoning, understanding and execution.
\newblock \emph{arXiv preprint arXiv:2401.03065}, 2024.

\bibitem[Lozhkov et~al.(2024)Lozhkov, Li, Allal, Cassano, Lamy-Poirier, Tazi, Tang, Pykhtar, Liu, Wei, Liu, Tian, Kocetkov, Zucker, Belkada, Wang, Liu, Abulkhanov, Paul, Li, Li, Risdal, Li, Zhu, Zhuo, Zheltonozhskii, Dade, Yu, Krauß, Jain, Su, He, Dey, Abati, Chai, Muennighoff, Tang, Oblokulov, Akiki, Marone, Mou, Mishra, Gu, Hui, Dao, Zebaze, Dehaene, Patry, Xu, McAuley, Hu, Scholak, Paquet, Robinson, Anderson, Chapados, Patwary, Tajbakhsh, Jernite, Ferrandis, Zhang, Hughes, Wolf, Guha, von Werra, and de~Vries]{lozhkov2024starcoder2stackv2}
Anton Lozhkov, Raymond Li, Loubna~Ben Allal, Federico Cassano, Joel Lamy-Poirier, Nouamane Tazi, Ao~Tang, Dmytro Pykhtar, Jiawei Liu, Yuxiang Wei, Tianyang Liu, Max Tian, Denis Kocetkov, Arthur Zucker, Younes Belkada, Zijian Wang, Qian Liu, Dmitry Abulkhanov, Indraneil Paul, Zhuang Li, Wen-Ding Li, Megan Risdal, Jia Li, Jian Zhu, Terry~Yue Zhuo, Evgenii Zheltonozhskii, Nii Osae~Osae Dade, Wenhao Yu, Lucas Krauß, Naman Jain, Yixuan Su, Xuanli He, Manan Dey, Edoardo Abati, Yekun Chai, Niklas Muennighoff, Xiangru Tang, Muhtasham Oblokulov, Christopher Akiki, Marc Marone, Chenghao Mou, Mayank Mishra, Alex Gu, Binyuan Hui, Tri Dao, Armel Zebaze, Olivier Dehaene, Nicolas Patry, Canwen Xu, Julian McAuley, Han Hu, Torsten Scholak, Sebastien Paquet, Jennifer Robinson, Carolyn~Jane Anderson, Nicolas Chapados, Mostofa Patwary, Nima Tajbakhsh, Yacine Jernite, Carlos~Muñoz Ferrandis, Lingming Zhang, Sean Hughes, Thomas Wolf, Arjun Guha, Leandro von Werra, and Harm de~Vries.
\newblock Starcoder 2 and the stack v2: The next generation, 2024.
\newblock URL \url{https://arxiv.org/abs/2402.19173}.

\bibitem[Ni et~al.(2024)Ni, Allamanis, Cohan, Deng, Shi, Sutton, and Yin]{ni2024nextteachinglargelanguage}
Ansong Ni, Miltiadis Allamanis, Arman Cohan, Yinlin Deng, Kensen Shi, Charles Sutton, and Pengcheng Yin.
\newblock Next: Teaching large language models to reason about code execution, 2024.
\newblock URL \url{https://arxiv.org/abs/2404.14662}.

\bibitem[Nye et~al.(2021)Nye, Andreassen, Gur{-}Ari, Michalewski, Austin, Bieber, Dohan, Lewkowycz, Bosma, Luan, Sutton, and Odena]{scratchpad}
Maxwell~I. Nye, Anders~Johan Andreassen, Guy Gur{-}Ari, Henryk Michalewski, Jacob Austin, David Bieber, David Dohan, Aitor Lewkowycz, Maarten Bosma, David Luan, Charles Sutton, and Augustus Odena.
\newblock Show your work: Scratchpads for intermediate computation with language models.
\newblock \emph{CoRR}, abs/2112.00114, 2021.
\newblock URL \url{https://arxiv.org/abs/2112.00114}.

\bibitem[OpenAI et~al.(2024)OpenAI, Achiam, Adler, Agarwal, Ahmad, Akkaya, Aleman, Almeida, Altenschmidt, Altman, Anadkat, Avila, Babuschkin, Balaji, Balcom, Baltescu, Bao, Bavarian, Belgum, Bello, Berdine, Bernadett-Shapiro, Berner, Bogdonoff, Boiko, Boyd, Brakman, Brockman, Brooks, Brundage, Button, Cai, Campbell, Cann, Carey, Carlson, Carmichael, Chan, Chang, Chantzis, Chen, Chen, Chen, Chen, Chen, Chess, Cho, Chu, Chung, Cummings, Currier, Dai, Decareaux, Degry, Deutsch, Deville, Dhar, Dohan, Dowling, Dunning, Ecoffet, Eleti, Eloundou, Farhi, Fedus, Felix, Fishman, Forte, Fulford, Gao, Georges, Gibson, Goel, Gogineni, Goh, Gontijo-Lopes, Gordon, Grafstein, Gray, Greene, Gross, Gu, Guo, Hallacy, Han, Harris, He, Heaton, Heidecke, Hesse, Hickey, Hickey, Hoeschele, Houghton, Hsu, Hu, Hu, Huizinga, Jain, Jain, Jang, Jiang, Jiang, Jin, Jin, Jomoto, Jonn, Jun, Kaftan, Łukasz Kaiser, Kamali, Kanitscheider, Keskar, Khan, Kilpatrick, Kim, Kim, Kim, Kirchner, Kiros, Knight, Kokotajlo, Łukasz Kondraciuk, Kondrich,
  Konstantinidis, Kosic, Krueger, Kuo, Lampe, Lan, Lee, Leike, Leung, Levy, Li, Lim, Lin, Lin, Litwin, Lopez, Lowe, Lue, Makanju, Malfacini, Manning, Markov, Markovski, Martin, Mayer, Mayne, McGrew, McKinney, McLeavey, McMillan, McNeil, Medina, Mehta, Menick, Metz, Mishchenko, Mishkin, Monaco, Morikawa, Mossing, Mu, Murati, Murk, Mély, Nair, Nakano, Nayak, Neelakantan, Ngo, Noh, Ouyang, O'Keefe, Pachocki, Paino, Palermo, Pantuliano, Parascandolo, Parish, Parparita, Passos, Pavlov, Peng, Perelman, de~Avila Belbute~Peres, Petrov, de~Oliveira~Pinto, Michael, Pokorny, Pokrass, Pong, Powell, Power, Power, Proehl, Puri, Radford, Rae, Ramesh, Raymond, Real, Rimbach, Ross, Rotsted, Roussez, Ryder, Saltarelli, Sanders, Santurkar, Sastry, Schmidt, Schnurr, Schulman, Selsam, Sheppard, Sherbakov, Shieh, Shoker, Shyam, Sidor, Sigler, Simens, Sitkin, Slama, Sohl, Sokolowsky, Song, Staudacher, Such, Summers, Sutskever, Tang, Tezak, Thompson, Tillet, Tootoonchian, Tseng, Tuggle, Turley, Tworek, Uribe, Vallone, Vijayvergiya,
  Voss, Wainwright, Wang, Wang, Wang, Ward, Wei, Weinmann, Welihinda, Welinder, Weng, Weng, Wiethoff, Willner, Winter, Wolrich, Wong, Workman, Wu, Wu, Wu, Xiao, Xu, Yoo, Yu, Yuan, Zaremba, Zellers, Zhang, Zhang, Zhao, Zheng, Zhuang, Zhuk, and Zoph]{openai2024gpt4technicalreport}
OpenAI, Josh Achiam, Steven Adler, Sandhini Agarwal, Lama Ahmad, Ilge Akkaya, Florencia~Leoni Aleman, Diogo Almeida, Janko Altenschmidt, Sam Altman, Shyamal Anadkat, Red Avila, Igor Babuschkin, Suchir Balaji, Valerie Balcom, Paul Baltescu, Haiming Bao, Mohammad Bavarian, Jeff Belgum, Irwan Bello, Jake Berdine, Gabriel Bernadett-Shapiro, Christopher Berner, Lenny Bogdonoff, Oleg Boiko, Madelaine Boyd, Anna-Luisa Brakman, Greg Brockman, Tim Brooks, Miles Brundage, Kevin Button, Trevor Cai, Rosie Campbell, Andrew Cann, Brittany Carey, Chelsea Carlson, Rory Carmichael, Brooke Chan, Che Chang, Fotis Chantzis, Derek Chen, Sully Chen, Ruby Chen, Jason Chen, Mark Chen, Ben Chess, Chester Cho, Casey Chu, Hyung~Won Chung, Dave Cummings, Jeremiah Currier, Yunxing Dai, Cory Decareaux, Thomas Degry, Noah Deutsch, Damien Deville, Arka Dhar, David Dohan, Steve Dowling, Sheila Dunning, Adrien Ecoffet, Atty Eleti, Tyna Eloundou, David Farhi, Liam Fedus, Niko Felix, Simón~Posada Fishman, Juston Forte, Isabella Fulford, Leo
  Gao, Elie Georges, Christian Gibson, Vik Goel, Tarun Gogineni, Gabriel Goh, Rapha Gontijo-Lopes, Jonathan Gordon, Morgan Grafstein, Scott Gray, Ryan Greene, Joshua Gross, Shixiang~Shane Gu, Yufei Guo, Chris Hallacy, Jesse Han, Jeff Harris, Yuchen He, Mike Heaton, Johannes Heidecke, Chris Hesse, Alan Hickey, Wade Hickey, Peter Hoeschele, Brandon Houghton, Kenny Hsu, Shengli Hu, Xin Hu, Joost Huizinga, Shantanu Jain, Shawn Jain, Joanne Jang, Angela Jiang, Roger Jiang, Haozhun Jin, Denny Jin, Shino Jomoto, Billie Jonn, Heewoo Jun, Tomer Kaftan, Łukasz Kaiser, Ali Kamali, Ingmar Kanitscheider, Nitish~Shirish Keskar, Tabarak Khan, Logan Kilpatrick, Jong~Wook Kim, Christina Kim, Yongjik Kim, Jan~Hendrik Kirchner, Jamie Kiros, Matt Knight, Daniel Kokotajlo, Łukasz Kondraciuk, Andrew Kondrich, Aris Konstantinidis, Kyle Kosic, Gretchen Krueger, Vishal Kuo, Michael Lampe, Ikai Lan, Teddy Lee, Jan Leike, Jade Leung, Daniel Levy, Chak~Ming Li, Rachel Lim, Molly Lin, Stephanie Lin, Mateusz Litwin, Theresa Lopez, Ryan
  Lowe, Patricia Lue, Anna Makanju, Kim Malfacini, Sam Manning, Todor Markov, Yaniv Markovski, Bianca Martin, Katie Mayer, Andrew Mayne, Bob McGrew, Scott~Mayer McKinney, Christine McLeavey, Paul McMillan, Jake McNeil, David Medina, Aalok Mehta, Jacob Menick, Luke Metz, Andrey Mishchenko, Pamela Mishkin, Vinnie Monaco, Evan Morikawa, Daniel Mossing, Tong Mu, Mira Murati, Oleg Murk, David Mély, Ashvin Nair, Reiichiro Nakano, Rajeev Nayak, Arvind Neelakantan, Richard Ngo, Hyeonwoo Noh, Long Ouyang, Cullen O'Keefe, Jakub Pachocki, Alex Paino, Joe Palermo, Ashley Pantuliano, Giambattista Parascandolo, Joel Parish, Emy Parparita, Alex Passos, Mikhail Pavlov, Andrew Peng, Adam Perelman, Filipe de~Avila Belbute~Peres, Michael Petrov, Henrique~Ponde de~Oliveira~Pinto, Michael, Pokorny, Michelle Pokrass, Vitchyr~H. Pong, Tolly Powell, Alethea Power, Boris Power, Elizabeth Proehl, Raul Puri, Alec Radford, Jack Rae, Aditya Ramesh, Cameron Raymond, Francis Real, Kendra Rimbach, Carl Ross, Bob Rotsted, Henri Roussez,
  Nick Ryder, Mario Saltarelli, Ted Sanders, Shibani Santurkar, Girish Sastry, Heather Schmidt, David Schnurr, John Schulman, Daniel Selsam, Kyla Sheppard, Toki Sherbakov, Jessica Shieh, Sarah Shoker, Pranav Shyam, Szymon Sidor, Eric Sigler, Maddie Simens, Jordan Sitkin, Katarina Slama, Ian Sohl, Benjamin Sokolowsky, Yang Song, Natalie Staudacher, Felipe~Petroski Such, Natalie Summers, Ilya Sutskever, Jie Tang, Nikolas Tezak, Madeleine~B. Thompson, Phil Tillet, Amin Tootoonchian, Elizabeth Tseng, Preston Tuggle, Nick Turley, Jerry Tworek, Juan Felipe~Cerón Uribe, Andrea Vallone, Arun Vijayvergiya, Chelsea Voss, Carroll Wainwright, Justin~Jay Wang, Alvin Wang, Ben Wang, Jonathan Ward, Jason Wei, CJ~Weinmann, Akila Welihinda, Peter Welinder, Jiayi Weng, Lilian Weng, Matt Wiethoff, Dave Willner, Clemens Winter, Samuel Wolrich, Hannah Wong, Lauren Workman, Sherwin Wu, Jeff Wu, Michael Wu, Kai Xiao, Tao Xu, Sarah Yoo, Kevin Yu, Qiming Yuan, Wojciech Zaremba, Rowan Zellers, Chong Zhang, Marvin Zhang, Shengjia
  Zhao, Tianhao Zheng, Juntang Zhuang, William Zhuk, and Barret Zoph.
\newblock Gpt-4 technical report, 2024.
\newblock URL \url{https://arxiv.org/abs/2303.08774}.

\bibitem[Rozière et~al.(2024)Rozière, Gehring, Gloeckle, Sootla, Gat, Tan, Adi, Liu, Sauvestre, Remez, Rapin, Kozhevnikov, Evtimov, Bitton, Bhatt, Ferrer, Grattafiori, Xiong, Défossez, Copet, Azhar, Touvron, Martin, Usunier, Scialom, and Synnaeve]{roziere2024codellamaopenfoundation}
Baptiste Rozière, Jonas Gehring, Fabian Gloeckle, Sten Sootla, Itai Gat, Xiaoqing~Ellen Tan, Yossi Adi, Jingyu Liu, Romain Sauvestre, Tal Remez, Jérémy Rapin, Artyom Kozhevnikov, Ivan Evtimov, Joanna Bitton, Manish Bhatt, Cristian~Canton Ferrer, Aaron Grattafiori, Wenhan Xiong, Alexandre Défossez, Jade Copet, Faisal Azhar, Hugo Touvron, Louis Martin, Nicolas Usunier, Thomas Scialom, and Gabriel Synnaeve.
\newblock Code llama: Open foundation models for code, 2024.
\newblock URL \url{https://arxiv.org/abs/2308.12950}.

\bibitem[Zaremba \& Sutskever(2014)Zaremba and Sutskever]{learning_to_execute2014}
Wojciech Zaremba and Ilya Sutskever.
\newblock Learning to execute.
\newblock \emph{CoRR}, abs/1410.4615, 2014.
\newblock URL \url{http://arxiv.org/abs/1410.4615}.

\end{thebibliography}
\bibliographystyle{iclr2025_conference}

\clearpage

\appendix

\section{Long Execution Fine-Grained Results}
\label{app:long_res_fine}

Table \ref{tab:collatz} provides the fine-grained results for the long executions.

\begin{table}[thbp!]

\caption{\label{tab:collatz}
Collatz, Fibonacci, and binary counter results. Reported accuracy of each execution final result and number of steps needed between parentheses.}

\scalebox{0.73}{
    \begin{tabular}{l|cccccccccc}
\hline
\textsc{Representation} & \multicolumn{10}{c}{\textsc{Collatz}} \\
& n=4 & n=5 & n=8 & n=18 & n=103 & n=457 & n=1127 & n=2620 & n=3038 & Acc. (avg steps) \\
\hline
Output & \checkmark (1) & \checkmark  (1) & \checkmark (1) & \checkmark (1) & $\times$ & $\times$ & $\times$ & $\times$ & $\times$ & 4/9 (1) \\
\hline

Scratchpad & \checkmark (13) & \checkmark (25) & \checkmark (17) & \checkmark (85) & $\times$ & $\times$ & $\times$ & $\times$ & $\times$ & 4/9 (35) \\
\hline
Compact scratchpad & \checkmark (13) & \checkmark (25) & \checkmark (17) & \checkmark (85) & \checkmark (353) & $\times$ & $\times$ & $\times$ & $\times$  & 5/9 (98.6) \\
Line-1 & \checkmark (11) & \checkmark (23) & \checkmark (15) & \checkmark (83) & \checkmark (351) & \checkmark (515) & \checkmark (551) & $\times$ & \checkmark (619) & 8/9 (271) \\
\hline
Line-n & \checkmark (2) & \checkmark (9)  & \checkmark (5)  & \checkmark (41)  & \checkmark (133) & \checkmark (199) & \checkmark (213) & \checkmark (247) & $\times$ & 8/9 (106.1) \\
Line-n + Dijsktra & \checkmark (2) & \checkmark (3) & \checkmark (2) & \checkmark (9) & \checkmark (36) & \checkmark (53) & \checkmark (57) & \checkmark (62) & \checkmark (67) & 9/9 (32.3) \\
\hline \hline

& \multicolumn{10}{c}{\textsc{Binary Counter}} \\
& n=4 & n=5 & n=8 & n=18 & n=103 & n=457 & n=1127 & n=2620 & n=3038 & Agg. \\
\hline
Output  & $\times$ & $\times$ & \checkmark (1) & $\times$ & $\times$ & $\times$ & $\times$ & $\times$ & $\times$ & 1/9 (1)\\
\hline

Scratchpad & \checkmark (24) & \checkmark (27) & \checkmark (43) & \checkmark (88) & $\times$ & $\times$ & $\times$ & $\times$ & $\times$ & 4/9 (45.5) \\
\hline
Compact scratchpad  & \checkmark (26) & \checkmark (27) & \checkmark (43) & \checkmark (88) & \checkmark (479) & $\times$ & $\times$ & $\times$ & $\times$ & 5/9 (116.2) \\
Line-1 & \checkmark (24) & \checkmark (27) & \checkmark (43) & \checkmark (88) & \checkmark (479) & \checkmark (2118) & \checkmark (5215) & $\times$ & \checkmark (14055) & 8/9 (2441) \\
\hline
Line-n & \checkmark (6) & $\times$ & $\times$ & $\times$ & $\times$ & $\times$ & $\times$ & $\times$ & $\times$ & 1/9 (6) \\
Line-n + Dijsktra &  \checkmark (3) &  \checkmark (3) &  \checkmark (5) &   \checkmark (10) & \checkmark (52) & \checkmark (229) & \checkmark (564) & \checkmark (1310) & \checkmark (1520) & 9/9 (410.7)\\
\hline \hline

& \multicolumn{10}{c}{\textsc{Fibonacci}} \\
& n=4 & n=5 & n=8 & n=18 & n=103 &  &  &  &  & Agg. \\
\hline
Output  & \checkmark (1) & \checkmark (1) & \checkmark (1) & \checkmark (1) & $\times$ &  &  &  &  & 4/5 (1) \\
\hline

Scratchpad & \checkmark (18) & \checkmark (22) & \checkmark (34) & \checkmark (74) & $\times$ &  &  &  &  & 4/5  (37)\\
\hline
Compact scratchpad & \checkmark (18) & \checkmark (22) & \checkmark (34) & \checkmark (74) & \checkmark (414) &  &  &  &  & 5/5 (140.5) \\
Line-1 & \checkmark (18) & \checkmark (22) & \checkmark (34) & \checkmark (74) & \checkmark (414) &  &  &  &  & 5/5 (140.5) \\
\hline
Line-n & \checkmark (11) & \checkmark (11) & \checkmark (15) & \checkmark (37) & \checkmark (160) &  &  &  &  & 5/5 (46.8) \\
Line-n + Dijsktra & \checkmark (2) & \checkmark (3) & \checkmark (4)& \checkmark (8) & \checkmark (42) &  &  &  &  & 5/5 (11.8)\\
\hline

\end{tabular}
}
\end{table}

\section{Additional information on long executions}

Here we provide the implementations of the algorithmic tasks used for the long executions section. Note that to encourage models to attend rather than memorized, in this case we replace function names with \texttt{f} when ingesting these functions to the models.

\subsection{Collatz}

\texttt{collatz} returns the number of iterations needed to arrive to 1 in the Collazt sequence.
\begin{verbatim}

def collatz(n):
    steps = 0
    while n > 1:
        steps += 1
        if n % 2 == 0:
            n = n // 2
        else:
            n = 3 * n + 1
    return steps

\end{verbatim}

\subsection{Binary counter}
\texttt{binary\_counter} implements a 4-bit binary counter by hand.

\begin{verbatim}
def binary_counter(n):
    a = False
    b = False
    c = False
    d = False
    for i in range(n):
        if not d:
            d = True
        elif not c:
            c = True
            d = False
        elif not b:
            b = True
            c = False
            d = False
        else:
            a = not a
            b = False
            c = False
            d = False
    return a, b, c, d
\end{verbatim}

\subsection{Iterative Fibonacci}
\texttt{fibonacci} is an iterative implementation of Fibonacci.

\begin{verbatim}
def fibonacci(n):
    if n == 0:
        return 0
    elif n == 1:
        return 1
    prev_prev = 0
    prev = 1
    for i in range(2, n + 1):
        curr = prev_prev + prev
        prev_prev = prev
        prev = curr
    return prev
\end{verbatim}

\clearpage

\section{Input Prediction Results}

\begin{figure}[ht]
    \centering
    \includegraphics[width=\textwidth]{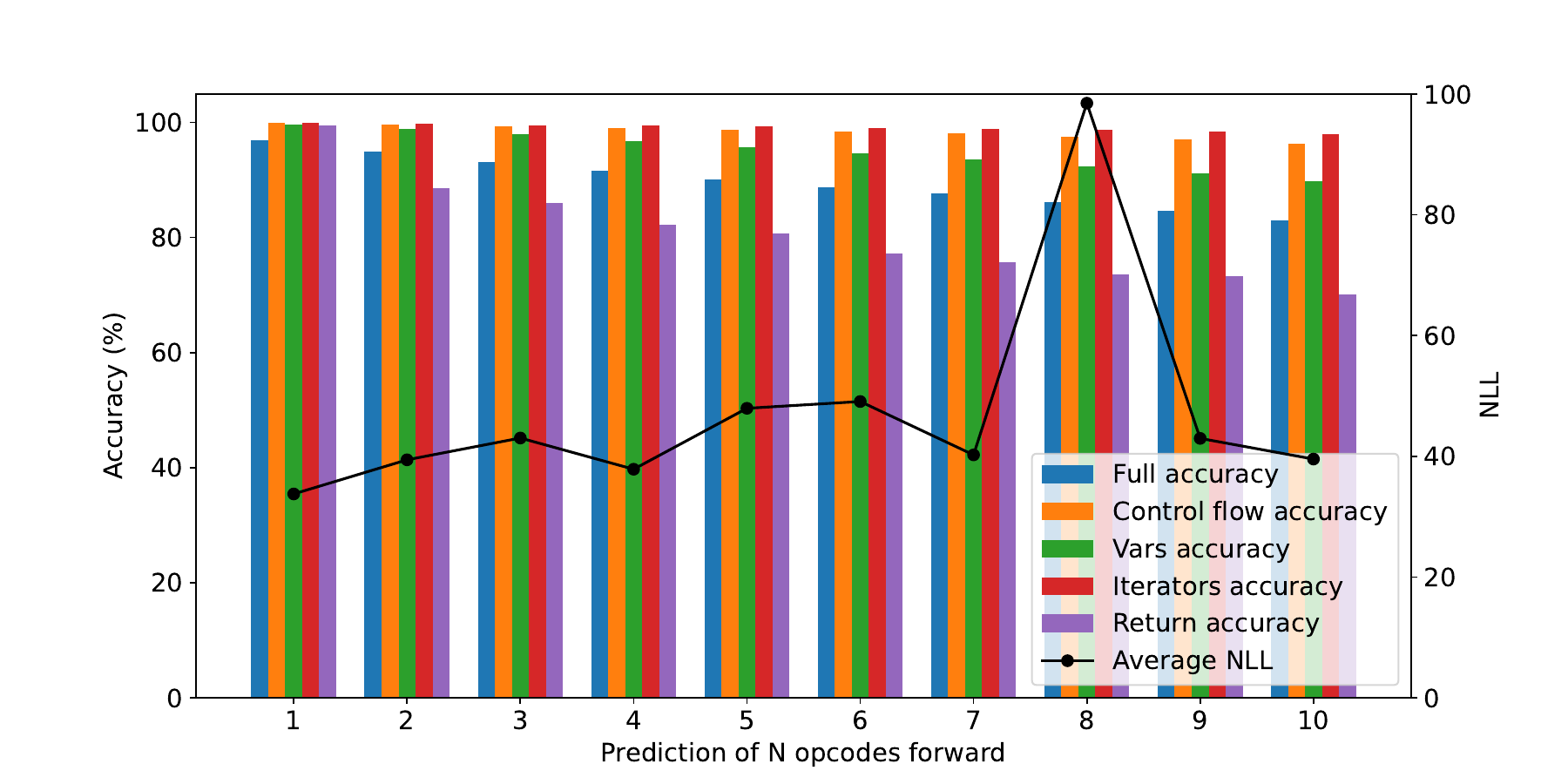}
    \caption{Plot showing individual state prediction performance when increasing N instructions into the future, compared to the predictions NLL. NLL stdev omitted for clarity.}
    \label{fig:variable_bc_ns}
\end{figure}

Table \ref{tab:intrinsic_crux_e_rev} and Figure \ref{fig:variable_bc_ns} show the results for Crux-I (i.e., input prediction given output prediction on CruxEval). We note that the reported results are strict lower bounds on the accuracy, given that multiple inputs are possible given the same output and we evaluated on exact match.
 
\begin{table}[ht]

\caption{\label{tab:intrinsic_crux_e_rev}
Individual trace evaluations on reversed CruxEval (i.e., predicting previous steps from future ones).
}
\centering
\scalebox{0.9}{
\begin{tabular}{lllllll}

\textsc{Representation} & \textsc{Model} & \textsc{Control flow} & \textsc{Vars}  & \textsc{Iterator}  & \textsc{Full}\\
\hline

\hline
line-1-rev  & Llama3.1 8B (Prompted) & 51.6\% & 38.9\% &  48\%& 11\% \\
& \hspace{3mm} + \name{} & 98.8\% & 88.4\%  & 99.6\%  &87.9\%\\

\hline 
line-n-rev (global) & Llama3.1 8B (Prompted) & 22.5\% & 13.6\% & 11.1\% & 1.6\% \\
 & \hspace{3mm} + \name{} & 94.1\% & 74\% & 93.3\% & 72.5\%\\

\hline

\hline
\end{tabular}
} 

\end{table}

\end{document}